\documentclass[10pt,twocolumn]{article}
\usepackage[a4paper,top=18mm,bottom=20mm,left=16mm,right=16mm,columnsep=7mm]{geometry}
\usepackage{amsmath}
\usepackage{newtxtext,newtxmath}
\usepackage{microtype}
\usepackage{xcolor}
\usepackage{graphicx}
\usepackage{booktabs}
\usepackage{tabularx}
\usepackage{array}
\usepackage{hyperref}
\usepackage{xurl}
\usepackage{titlesec}
\usepackage{enumitem}
\usepackage{caption}
\usepackage{float}
\usepackage{placeins}
\usepackage{needspace}
\usepackage{balance}
\definecolor{vlteal}{HTML}{16615A}
\definecolor{vltealdark}{HTML}{0E4742}
\hypersetup{colorlinks=true,linkcolor=vltealdark,citecolor=vltealdark,urlcolor=vlteal,
  pdftitle={Engineering Efficient Self-Play Chess},pdfauthor={Bertil Braun}}
\titleformat{\section}{\large\bfseries}{\thesection}{0.55em}{}
\titleformat{\subsection}{\normalsize\bfseries}{\thesubsection}{0.5em}{}
\titlespacing*{\section}{0pt}{1.15em}{0.45em}
\titlespacing*{\subsection}{0pt}{0.9em}{0.3em}
\setlist[itemize]{leftmargin=1.25em,itemsep=0.08em,topsep=0.25em}
\setlist[enumerate]{leftmargin=1.35em,itemsep=0.12em,topsep=0.25em}
\title{\textbf{Engineering Efficient Self-Play Chess: Search, Replay, and\\Throughput Under Limited Compute}}
\author{Bertil Braun\\
  \small \href{mailto:contact@bertil-braun.de}{contact@bertil-braun.de}}
\date{}
\begin{document}
\twocolumn[{
\begin{@twocolumnfalse}
\maketitle
\vspace{-1.8em}
\begin{abstract}
How strong can an AlphaZero-style chess system become under limited training compute when its entire learning loop is engineered for efficiency? We train from random initialization through searched self-play on a single eight-GPU node for 2.5 days. The resulting 6.32-million-parameter model reaches 3,251 benchmark Elo [3,206, 3,297] at 100,000 searches per move (estimated at under five seconds of thinking time) against a fixed-node Stockfish 13 ladder. The run ingests 3.25 million completed games, involves an estimated 100 billion search simulations, and makes 836.6 million training presentations. We investigate search allocation, replay and restart-state selection, policy representation, progressive model sizing, quantized inference, and throughput engineering. Alongside the retained design, we document plausible alternatives that failed to improve the complete learning loop or did not justify their cost. The reported strength is a result of the integrated system, not an isolated Elo gain attributable to any single choice.
\end{abstract}
\vspace{0.45em}
\vspace{1.0em}
\end{@twocolumnfalse}
}]
\section{How far can efficient self-play go?}\label{sec:01-motivation-and-scope}
An AlphaZero-style chess player \cite{ref1} improves by searching its own games, learning from the resulting positions, and repeating that cycle with a stronger network. Search creates targets; replay decides which targets persist; training absorbs them; and evaluation must distinguish progress from noise. Under a limited budget, improving any one step matters only if the complete loop produces stronger play within the available time.

This study asks how strong that loop can become on a single eight-GPU node when the system is engineered for efficiency. Over \textbf{2.5 days} of training from random weights on searched self-play games, the run produced a 6.32-million-parameter model. It reached \textbf{1,658 benchmark Elo without search} and \textbf{3,251 benchmark Elo at 100,000 searches per move}---estimated at under five seconds of thinking time---against the project's fixed-node Stockfish 13 ladder. Table \ref{tab:02-methodology-and-evidence-1} gives the match results; Chapter \ref{sec:07-final-run-results} follows the model's improvement across search budgets.

The result depends on the whole learning loop. The network's representation affects both what it can learn and how quickly it can supply search. Native search, batched inference, and TensorRT make enough searched games available for training; replay and training must then turn those games into stronger play. Several plausible optimizations improved a local metric without improving that complete loop.

\subsection{Scope and contributions}\label{sec:01-motivation-and-scope-scope-and-contributions}
Chess is the subject of the study. The runtime also supports Go, and KataGo's fast and full searches inspired one of the approaches tested here \cite{ref2}. We briefly considered 7\ensuremath{\times}7 and 9\ensuremath{\times}9 Go as cheaper settings for tuning chess hyperparameters. In those exploratory games, we observed a strong first-player advantage, short trajectories, and a value target that learned quickly. Useful Go tuning appeared unlikely to transfer directly to chess, so we kept chess as the focus.

We examine the choices that shaped the final recipe: search allocation, graph search and caching, policy representation, model sizing, replay and restart states, resignation, auxiliary targets, and quantized inference. The report also traces the throughput needed to supply searched games and studies three failures with transferable lessons about self-play targets, deployment fidelity, and model promotion.

A live chess demonstration is also available \cite{ref12}.

\subsection{Roadmap}\label{sec:01-motivation-and-scope-roadmap}
Chapter \ref{sec:02-methodology-and-evidence} presents the chess evaluation. Chapter \ref{sec:03-system-and-methods} explains the AlphaZero learning principle and the system that implements it. Chapter \ref{sec:04-research-investigations} examines search, replay, and network design, followed by throughput engineering in Chapter \ref{sec:05-systems-optimization}. The three failure studies in Chapter \ref{sec:05a-three-failures} lead into training progress in Chapter \ref{sec:06-final-chess-recipe} and playing strength in Chapter \ref{sec:07-final-run-results}. Appendix \ref{app:D} specifies the final recipe.

\section{Evaluating the chess system}\label{sec:02-methodology-and-evidence}
We measure thinking effort in search visits rather than seconds. For a given model and search configuration, this budget is hardware-independent: faster hardware finishes sooner instead of receiving more search. Fixed budgets also keep background load from changing the amount of search performed. We report approximate thinking time separately to give these budgets a practical scale.

The most direct test of the system is whether its final model wins games. We played it against Stockfish 13 at fixed search limits, starting from 50 openings and playing each once with each colour. The model played without search and at four progressively larger search budgets. For each budget, we tested two Stockfish limits rather than trusting a single opponent.

\begin{table*}[!t]
\centering\small
\caption{Final checkpoint against fixed-node Stockfish 13}\label{tab:02-methodology-and-evidence-1}
\begin{tabular}{@{}llllll@{}}
\toprule
Model searches & Parallel & Stockfish nodes & W/D/L & Score & Benchmark Elo (95\% CI) \\
\midrule
Policy only & -- & 1,000 & 32/24/44 & 0.440 & \textbf{1,658 [1,608, 1,710]} \\
Policy only & -- & 2,000 & 16/22/62 & 0.270 & 1,717 [1,638, 1,790] \\
100 & 1 & 5,000 & 51/28/21 & 0.650 & 2,328 [2,276, 2,384] \\
100 & 1 & 10,000 & 39/18/43 & 0.480 & \textbf{2,456 [2,400, 2,512]} \\
1,000 & 1 & 20,000 & 47/40/13 & 0.670 & 2,823 [2,774, 2,873] \\
1,000 & 1 & 50,000 & 21/48/31 & 0.450 & \textbf{2,925 [2,875, 2,977]} \\
10,000 & 4 & 50,000 & 45/40/15 & 0.650 & 3,068 [3,023, 3,120] \\
10,000 & 4 & 100,000 & 30/44/26 & 0.520 & \textbf{3,114 [3,065, 3,163]} \\
100,000 & 16 & 100,000 & 51/38/11 & 0.700 & 3,247 [3,192, 3,305] \\
100,000 & 16 & 200,000 & 25/56/19 & 0.530 & \textbf{3,251 [3,206, 3,297]} \\
\bottomrule
\end{tabular}
\end{table*}
\subsection{Final playing strength}\label{sec:02-methodology-and-evidence-final-playing-strength}
Table \ref{tab:02-methodology-and-evidence-1} gives all ten matches. Each row contains 100 games; wins, draws, and losses are from our model's perspective. The bold rating for each model budget comes from the opponent against which it scored closest to 50\%, where the rating estimate needs the least extrapolation.

At the deepest budget, the model scored 3,251 benchmark Elo against the harder opponent and 3,247 against the other: the two measurements agree closely. The ratings use a published calibration of Stockfish's fixed node limits \cite{ref9}. Chapter \ref{sec:07-final-run-results} examines how strength changes with search; Appendix \ref{app:B} gives the rating calculation and interval method.

\subsection{Reading the component experiments}\label{sec:02-methodology-and-evidence-reading-the-component-experiments}
The rest of the paper asks why the system reached that result. A faster inference engine can supply more search, but that only helps training if the system finishes more games and admits useful positions to replay. A lower loss on stored positions can indicate a better fit, but the real test is whether the next model plays better. We follow each proposed improvement as far through this chain as the experiment measured.

For online comparisons, we matched starting weights, replay, evaluation, and hardware where possible. Smaller frozen-data tests helped screen ideas before expensive self-play runs. Because the final recipe combines many changes, we attribute a separate strength gain to a component only when a comparison actually measured one.

\section{Background and system design}\label{sec:03-system-and-methods}
This chapter introduces the AlphaZero learning loop and the system used to run it under limited compute. It follows the path from network predictions through search and self-play to replay and training, establishing the foundations for the experiments in Chapter \ref{sec:04-research-investigations}.

\subsection{Learning through search}\label{sec:03-system-and-methods-learning-through-search}
AlphaZero learns a chess player without examples of human play \cite{ref1}. Its network has two jobs: the \emph{policy} assigns probabilities to moves, while the \emph{value} estimates the outcome of a position. Monte Carlo tree search (MCTS) brings these predictions together. It explores moves using both their policy probabilities and the results found so far, balancing promising continuations against less-explored alternatives. At a new leaf, the network evaluates the position; that value is backed up along the path to inform subsequent exploration.

Search can therefore challenge the network's first impression. A promising move may reveal a strong reply for the opponent, while a less obvious move may lead to better positions. The root visit distribution becomes a policy target, and the completed game's outcome teaches the value prediction. Training folds this experience back into the network. Better predictions then guide later searches towards more useful continuations, sustaining a cycle of search, self-play, and learning.

This is the central opportunity under limited compute: spend search to discover improvements, then learn enough from those improvements that the next search starts from a stronger player. Cheap searches that mostly repeat the network's initial preference may produce many positions but little new policy information. Very deep searches can make good targets too expensive to supply in sufficient quantity. The system must make both searching and learning affordable.

\subsection{Related work}\label{sec:03-system-and-methods-related-work}
AlphaZero establishes the self-play learning framework \cite{ref1}; KataGo shows how substantially its compute requirements can be reduced through changes to search, training, and network architecture \cite{ref2}. KataGo is the closest practical precedent for this study's efficiency focus. Its fast/full search schedule and auxiliary objectives \cite{ref2}, along with later self-play methods \cite{ref7}, motivated several investigations here. Their usefulness still depends on the game: completing more long Go games and supplying more searched chess positions need not favour the same allocation of compute.

Several narrower lines of work address where that compute should go. Dynamic simulation MCTS studies when to stop search \cite{ref3}, and targeted search control starts self-play from archived states to explore beyond ordinary opening trajectories \cite{ref5}. Prioritized experience replay changes which stored examples are learned from again \cite{ref4}, while Monte Carlo graph search shares work across paths reaching the same state \cite{ref6}. These ideas motivate the allocation, replay, restart, and reuse experiments. This chapter explains the resulting system; Chapter \ref{sec:04-research-investigations} examines how the alternatives performed, and Chapter \ref{sec:05-systems-optimization} measures the throughput needed to run it.

\subsection{One learning cycle}\label{sec:03-system-and-methods-one-learning-cycle}
\begin{figure*}[t]
\centering
\includegraphics[width=494.476bp,height=0.70\textheight,keepaspectratio]{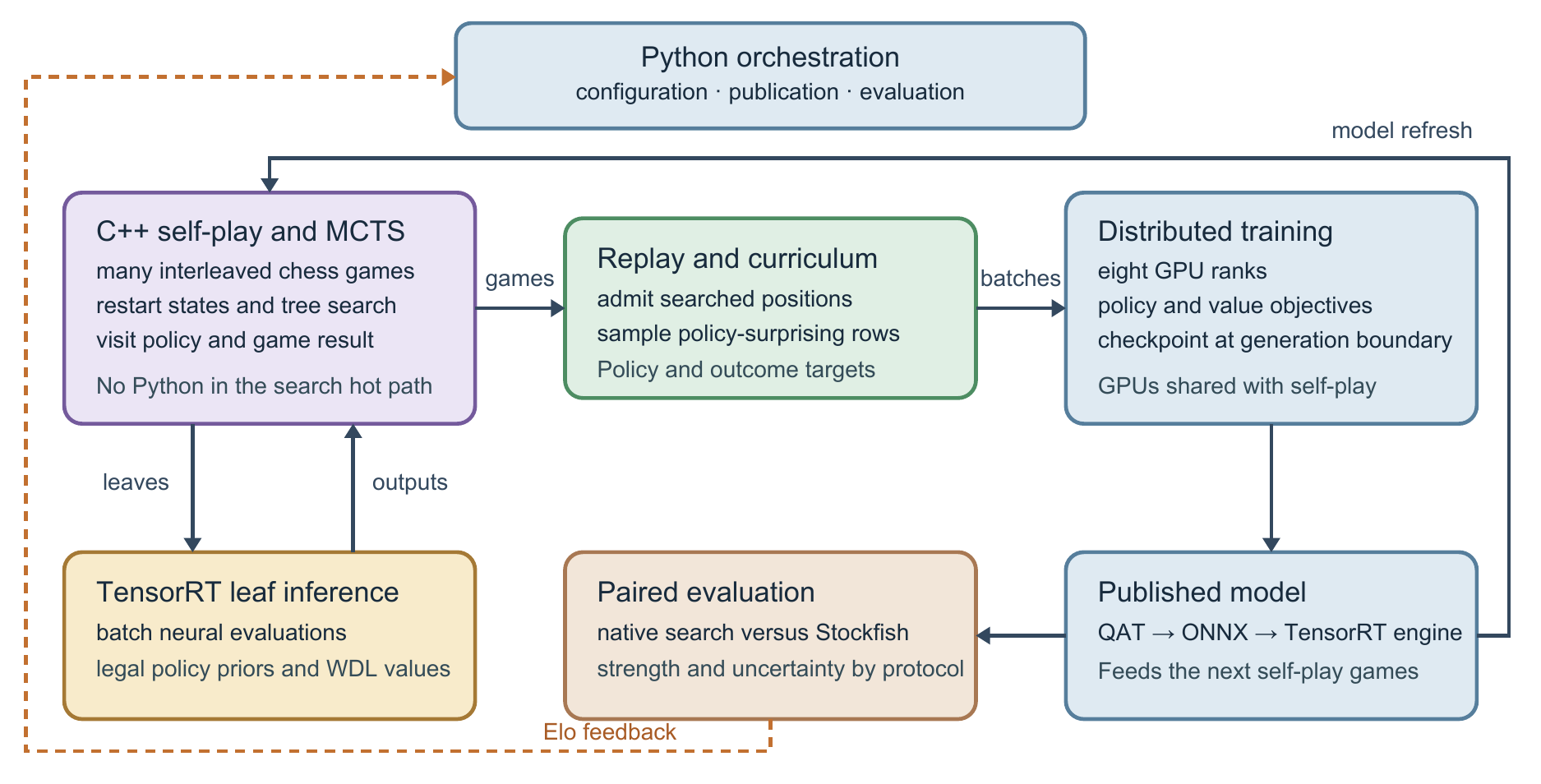}
\caption{Self-play turns a published network into searched games, replay supplies their positions to the trainer, and updated weights return to the actors. Separate matches measure the player's progress.}\label{fig:learning-loop}
\end{figure*}
The learning loop has three sources of work: actors play games, the replay pipeline stores their experience, and the trainer updates the network. An \emph{actor} is a self-play process that advances many games concurrently rather than waiting for each game to finish before starting another. All those games use a published set of network weights to guide their searches.

Figure \ref{fig:learning-loop} follows the flow between these components. Python orchestration starts and coordinates the work. Within each actor, native C++ search requests policy and value predictions for newly explored positions. Requests from different games are grouped into batches and evaluated on the GPU by TensorRT, the optimized inference engine. While a batch is being evaluated, other ready games can advance. Returned predictions let the waiting searches update their trees and eventually choose their moves.

A finished game supplies a sequence of positions, the search policy at each recorded move, and its final outcome. The replay pipeline converts that sequence into training examples. The trainer draws batches from the accumulated examples and adjusts the weights to better predict their policies and outcomes. After a block of optimizer steps, publication prepares the updated network for the actors. The next games then benefit from what the learner has absorbed.

Some actors keep playing while training runs, so the system can produce the next games while learning from earlier ones. Evaluation runs alongside this loop using the native chess engine to play matches against Stockfish. These matches measure progress; they do not become self-play training data.

\begin{figure*}[t]
\centering
\includegraphics[width=494.476bp,height=0.70\textheight,keepaspectratio]{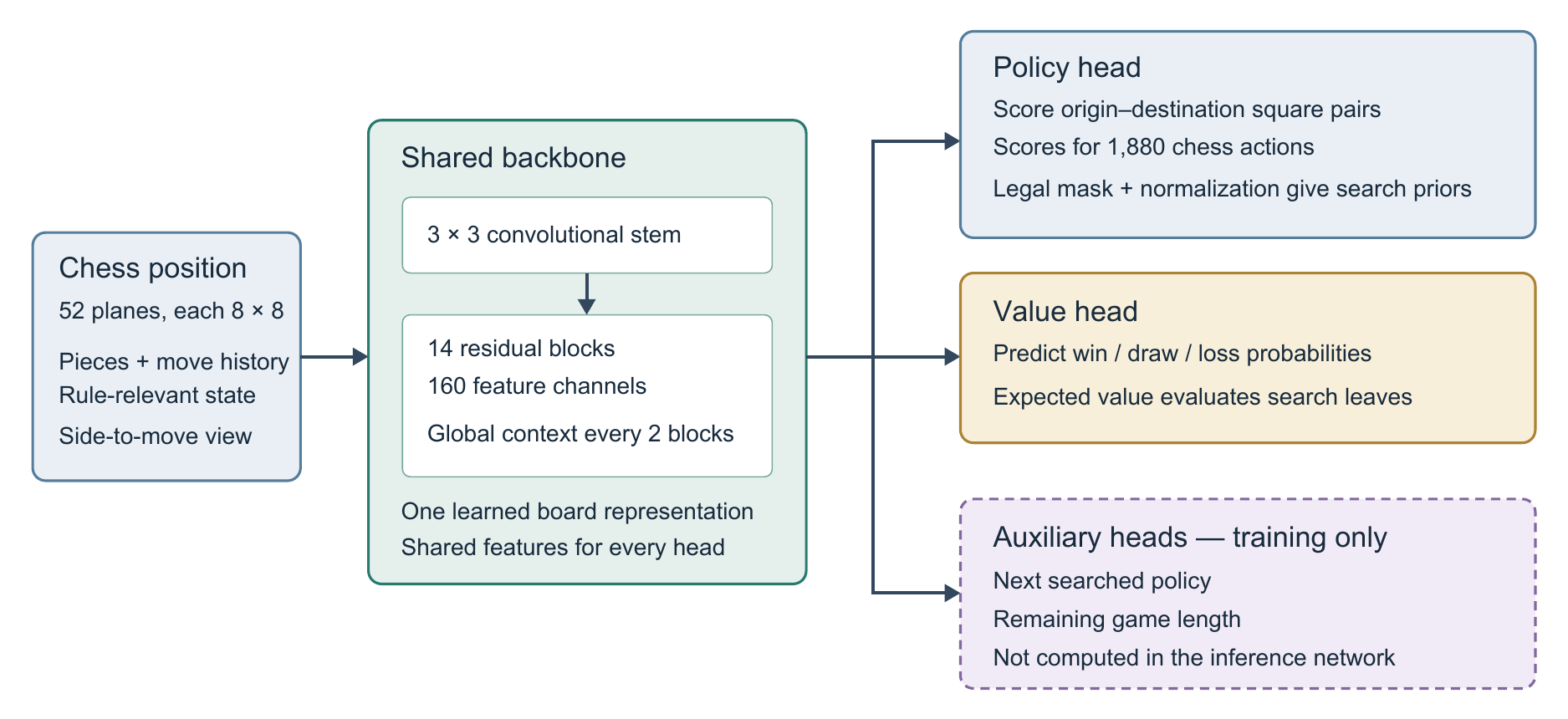}
\caption{The final chess network. A shared 14-block, 160-channel backbone processes 52 input planes. Policy and value heads provide the predictions used by search; the two auxiliary heads contribute only during training.}\label{fig:chess-network-architecture}
\end{figure*}
\subsection{Why the search loop stays in C++}\label{sec:03-system-and-methods-why-the-search-loop-stays-in-c}
The language boundary follows the frequency of the work. One played move requires hundreds or thousands of tree traversals. Each traversal may update a board, generate legal moves, select a continuation, encode a position, and back up a value. If Python dispatches these steps individually, interpreter and boundary-crossing costs recur throughout the most frequently executed part of the system. Even a fast GPU can then spend time waiting for the CPU to prepare its next batch.

C++ owns the complete inner loop: game state, search trees, traversal, and the inference request pipeline. An actor can retain its trees and buffers, interleave independent games, and submit batches without returning to Python for each search leaf. After a move is played, the relevant subtree becomes the next root, preserving useful work rather than rebuilding the search from nothing.

Python operates at a coarser scale. It coordinates processes, manages replay and publication, and trains the model through PyTorch and its distributed libraries. These tasks benefit from the existing numerical ecosystem and ease of experimentation without placing Python on the path of every simulation. The purpose of the split is not to make every operation native: it is to keep the expensive repeated work fast enough to supply the learner with games.

\subsection{Chess representation and outputs}\label{sec:03-system-and-methods-chess-representation-and-outputs}
Figure \ref{fig:chess-network-architecture} shows the network from its 52\ensuremath{\times}8\ensuremath{\times}8 chess input through a shared residual backbone to its output heads. The input planes encode pieces, recent moves, and game-state features from the side-to-move perspective. Castling rights, repetition, and the halfmove clock distinguish positions whose boards look identical but whose legal continuations or draw conditions differ. Appendix \ref{app:D}, Table \ref{tab:appendix-d-reproducibility-1} gives the complete channel-by-channel specification.

The backbone learns a common spatial representation for move selection and position evaluation. Its 14 residual blocks, each with 160 channels, refine board features by adding learned corrections to the preceding representation. Every second block also pools features across the board and feeds the resulting global context back into the spatial features. This lets a local feature respond to the wider position without relying solely on successive local convolutions to propagate that information.

A \emph{head} maps these shared features to a particular prediction. The policy head represents each square with learned origin and destination vectors; their dot products score moves, with additional offsets distinguishing promotion pieces. The fixed encoding contains 1,880 actions: 1,792 origin--destination pairs along sliding-piece rays or knight jumps, plus 88 promotion moves distinguishing the four promotion pieces. King and ordinary pawn moves use the same ray pairs. Mapping the scores to these actions, masking illegal moves, and normalizing produces the move probabilities that guide search. The value head instead predicts win, draw, and loss (WDL). Search uses the win probability minus the loss probability as its scalar estimate; training retains the full distribution, distinguishing a likely draw from equally likely winning and losing outcomes.

Both objectives train the same backbone, sharing the cost of extracting board features. Two auxiliary heads add supervision for the next searched policy and remaining game length. Their predictions are not needed to choose moves, so they are omitted from the inference copy while the backbone retains the features learned from those tasks. Appendix \ref{app:D} specifies the output shapes and action encoding.

\subsection{Search and self-play}\label{sec:03-system-and-methods-search-and-self-play}
At each move, search builds a tree rooted at the current position. Each traversal follows the PUCT selection rule, balancing the value found for a continuation against the benefit of investigating it further:

\begin{equation}
\begin{aligned}
a^* &= \arg\max_a \left[ Q(s,a) + U(s,a) \right],\\
U(s,a) &= c_{\mathrm{puct}} P(s,a)
\frac{\sqrt{\max(1,N(s))}}{1+N(s,a)}.
\end{aligned}
\end{equation}
Here, \emph{P(s,a)} is the policy prior for move \emph{a} in position \emph{s}, \emph{Q(s,a)} its estimated value for the player to move, and \emph{N(s,a)} its visit count; \emph{N(s)} counts visits to the parent position. The exploration constant is 1.5. The first term rewards continuations that search already considers strong. The second favours moves with a promising prior but relatively few visits, decaying as they receive attention. Unvisited moves start from a reduced parent-value estimate until a traversal supplies their own evidence.

Reaching a new leaf triggers a network evaluation, unless the game has ended and its result is already known. Backing up that value updates the visits and accumulated estimates along the selected path, reversing perspective at each ply. Successive traversals therefore test the network's preferences against increasingly explored replies, rather than merely resampling its initial move probabilities.

For example, the network may initially prefer a move that wins a pawn. Search can discover that the opponent then has a dangerous reply, lower its estimate of that continuation, and spend more visits on a safer alternative. The training signal is not just which move was finally played: the distribution of root visits records the search's relative preference across the available moves. That distribution becomes the policy target.

Self-play must explore as well as exploit its current knowledge. Noise added to the root prior makes games consider different continuations, and a temperature setting controls how sharply played moves follow the visit leader. The search budget rises in stages from 300 to 800 visits as training progresses. Useful visits from the retained subtree contribute to the next move, while each new search refines the choice from its new root.

Many independent games share inference batches. This uses the GPU efficiently without requiring every tree to search multiple leaves at once. When per-tree parallelism is used, several traversals can be waiting for results simultaneously; temporary reservations discourage them from all choosing the same path. They still make their choices without the other pending results, creating the strength-versus-latency tradeoff examined later.

Games need not all repeat the ordinary initial position. Roughly half start after a short random legal opening, and the other half use archived self-play positions with interesting alternatives left to explore. A restart plays a different branch from such a position, creating new experience rather than simply replaying its old target. Ordinary starts maintain whole-game coverage while restarts spend some of the budget on unresolved choices.

Games normally supply win, draw, or loss labels when they end. Resignation avoids spending search on clearly lost positions, but some games are forced to continue so the system can estimate how often resignation would be wrong. A maximum game length also prevents extremely long games from consuming unlimited compute. At that limit, one final full search estimates the unfinished position's value instead of assigning it an arbitrary result.

\subsection{Replay and materialization}\label{sec:03-system-and-methods-replay-and-materialization}
A \emph{replay buffer} separates the order in which experience is generated from the order in which it is learned. Adjacent positions in a game are highly correlated; mixing examples from many games gives each optimizer batch a broader range of openings, middlegames, and endings. Reuse also amortizes the cost of search: a useful position can contribute to several updates without requiring another game to generate it again.

Each example stores the encoded position, its legal actions, the searched policy, and an outcome target. For a finished game, the outcome is expressed from the side to move at each stored position. For a capped game, the final searched estimate supplies a substitute target. Random opening moves and reconstructed restart prefixes have no search policy of their own, so they are not treated as searched training examples.

\emph{Materialization} is the conversion from complete game trajectories to these rows. The rows live in a circular memory-mapped store: as the active window fills, new examples replace its oldest contents, and trainers read batches without loading the whole buffer into each process. Search policies are stored sparsely, retaining the actions with visit mass rather than writing a full mostly empty vector for every position.

Replay capacity grows from 600,000 towards 20 million positions. The small early window lets new experience quickly replace the weakest early play; the larger later window preserves more variety. Sampling combines a uniform component with a preference for positions where search changed the network's move probabilities substantially. Such positions offer a plausible learning opportunity, while uniform sampling prevents the learner from seeing only unusual or difficult cases. Each selected row has the same loss weight; priority changes how often it is seen.

Data supply also sets the pace of training. The chosen reuse ratio permits four training presentations per new replay position. A 500-step block with a batch of 2,048 therefore requires 256,000 new positions to support its 1,024,000 presentations. If the actors have not supplied enough, training waits. This keeps a faster optimizer from merely making more passes over an unchanged pool of experience.

\subsection{Training}\label{sec:03-system-and-methods-training}
Training distils search and game outcomes into network predictions. For one position, the objective is:

\begin{equation}
\begin{aligned}
\mathcal{L} &= \operatorname{CE}(\pi,p) + \operatorname{CE}(z,v)
+\mathcal{L}_{\mathrm{aux}}\\
&= -\sum_a \pi_a\log p_a
-\sum_{k\in\{W,D,L\}} z_k\log v_k
+\mathcal{L}_{\mathrm{aux}}.
\end{aligned}
\end{equation}
The searched policy \emph{\ensuremath{\pi}} and outcome target \emph{z} supervise the predicted move probabilities \emph{p} and WDL probabilities \emph{v}. Both primary losses have unit weight; the auxiliary term combines next-policy and remaining-length losses with weights 0.15 and 0.1. Their gradients meet in the shared backbone, so a feature useful for predicting outcomes can also improve the representation from which move preferences are learned.

Outcome discounting softens targets for positions far from the end of a game. It was introduced, alongside search discounting, to favour earlier conversion of winning positions; Section \ref{sec:04b-data-and-replay} discusses the motivation and unresolved benefit. A small contribution from the position's searched value is also blended into the training target. The auxiliary losses add next-policy and remaining-length supervision when those labels exist. An unfinished game cannot reveal its true remaining length, so that auxiliary loss is omitted rather than trained towards a fabricated zero.

Training uses distributed data parallel across all eight GPUs with a global batch size of 2,048.

Updates are grouped into blocks of 500 optimizer steps. A block followed by publication is a \emph{generation}, and a saved set of model weights is a \emph{checkpoint}. Half the self-play actors continue working while a block runs. This overlap is valuable because game production and training use the shared hardware differently, but also means that saving search work will shorten the full cycle only when search was delaying the next block.

\subsection{Progressive models and publication}\label{sec:03-system-and-methods-progressive-models-and-publication}
Starting with the final model size would spend its full inference cost even while the player is still learning basic chess. Progressive sizing starts with a smaller, faster network. Its searches are cheaper, supplying more games during this early stage. As improvement slows and model capacity becomes more limiting, a larger network can make better use of the experience already collected.

The ordinary transition trains a larger candidate on the same replay alongside the active model. Since the candidate starts independently, it first needs catch-up training before it can replace an already competent player. Paired head-to-head matches test whether it has caught up. These matches decide the occasional change in model size; they are not a gate applied to each routine training update. Within an active size, updated weights are published after training blocks without requiring them to defeat the previous generation.

A second approach initializes the larger network to reproduce the smaller network's current predictions. This function-preserving growth avoids much of the initial catch-up, but may also keep learning close to the smaller model's internal representation. The final reported player uses the successful small-to-medium path; the limited larger-model continuation did not establish a further strength gain.

Publication bridges training and inference. Training retains optimizer state and auxiliary heads, while search needs only a fast policy-and-value predictor. A separate inference copy removes those unused heads and combines operations where possible. It is exported through ONNX, a model representation that TensorRT can compile into an optimized GPU engine. Refitting replaces weights in a prepared engine rather than rebuilding it after every training block.

Most backbone convolutions use eight-bit integer arithmetic to accelerate inference. Quantization-aware training simulates the rounding and clipping this introduces, letting the network adapt before deployment. The exported copy is checked against the training model on representative positions so the speedup does not silently change its move preferences or value predictions. Once prepared, it becomes the model used by subsequent self-play.

\subsection{Evaluation}\label{sec:03-system-and-methods-evaluation}
Training loss shows how well a network fits replay, but fitting those examples is not the same as playing stronger chess. Evaluation therefore plays separate matches against fixed-node Stockfish opponents. Each opening is played twice with colours reversed, reducing the influence of which side received the easier starting position. Wins, draws, and losses determine the match score and its implied rating on the project's calibrated ladder.

During training, short 64-search and policy-only evaluations track progress at modest cost. They answer different questions: policy-only play tests the network's immediate move choice, while searched play tests how useful its policy and value are together in a tree. Larger final matches repeat the assessment across increasing search budgets, as reported in Table \ref{tab:02-methodology-and-evidence-1}.

The resulting feedback completes the experimental loop. A change that improves a local loss or throughput measure still has to help produce a stronger player within the available training time. The following chapters use this system description to examine those choices, rather than treating local speed or fit as an end in itself.

\section{Research investigations}\label{sec:research}\label{sec:04-research-investigations}
This chapter examines the approaches we tested, their results, and the choices retained in the chess system. Section \ref{sec:04a-search} concerns how much search to spend and where; Section \ref{sec:04b-data-and-replay} examines self-play data, restart states, and replay; Section \ref{sec:04c-networks-and-training} considers network architecture and training. Together, these investigations connect the cost of generating experience to how effectively the network learns from it.

\subsection{Spending search where it matters}\label{sec:04a-search}
More search makes an individual move stronger. With the same frozen network and opponent, raising the budget from 200 to 1,600 visits raised its match score from 0.318 to 0.748. Training poses a different question: could some moves use fewer visits so that self-play finishes more games without teaching the next model worse policies? Cheap moves might save computation, but the saved work matters only if it increases useful training data or shortens the learning cycle.

We tried fast/full searches, several ways to allocate or stop search according to position, and reuse through a graph or inference cache. The retained approach increases a fixed visit cap in stages and applies that cap to every recorded move. The experiments below distinguish four tests that can give different answers: immediate move strength, agreement with a deeper policy, strength of the next trained model, and throughput.

\begin{figure*}[t]
\centering
\includegraphics[width=494.476bp,height=0.70\textheight,keepaspectratio]{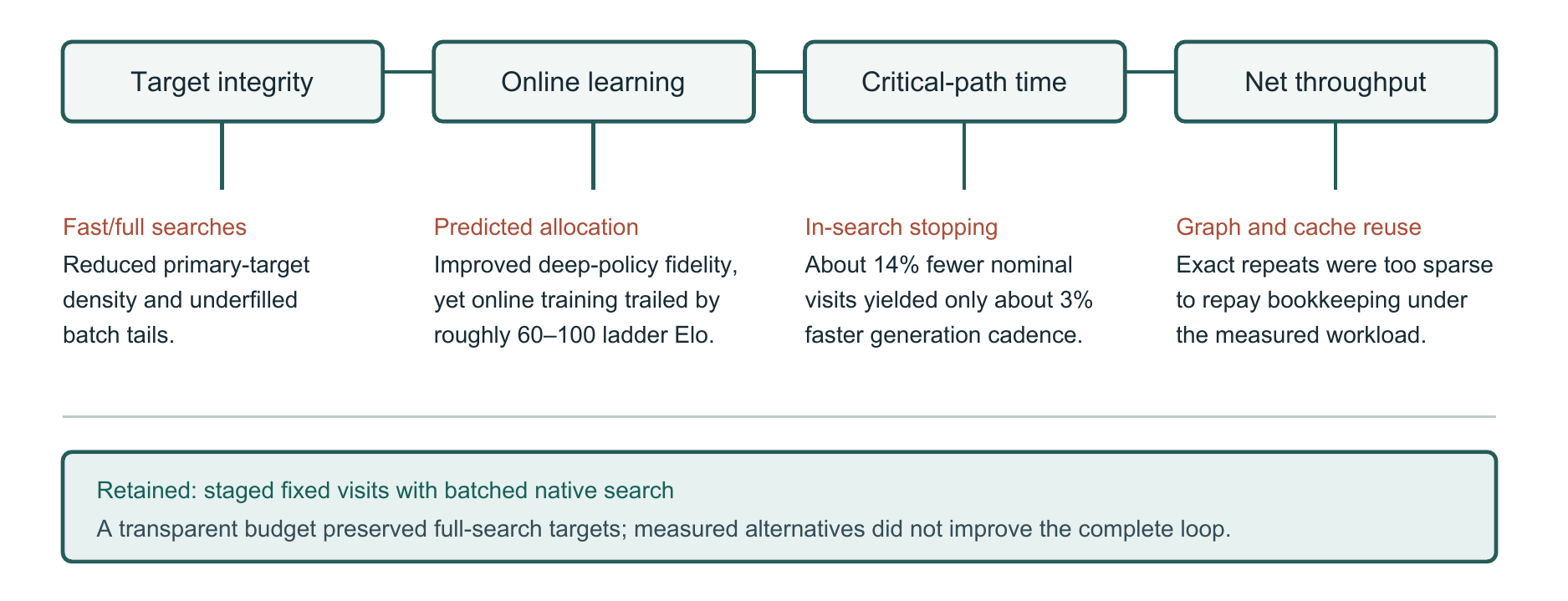}
\caption{Why the search alternatives were not retained. Fast/full search discarded useful targets; predicted allocation improved policy agreement but weakened learning; stopping saved search without much faster training; and exact graph or cache reuse saved too little work to repay its overhead.}\label{fig:search-decision-gates}
\end{figure*}
\subsubsection{The fixed-budget baseline}\label{sec:04a-search-the-fixed-budget-baseline}
The baseline uses the PUCT search described in Section \ref{sec:03-system-and-methods-search-and-self-play}, changing only the visit budget in the fixed-network comparison below.

Against the same opponent, score rose from 0.318 at 200 visits to 0.537, 0.580, 0.662, and 0.748 at 400, 600, 1,000, and 1,600 visits. Across the better-resolved middle and deep part of the sweep, the fitted trend was about 81 Elo per doubling. Individual adjacent comparisons were noisy, so the overall trend is more credible than any single increment. Search was still changing its policy target too: at 600 visits, only 72.4\% of positions selected the same best move as the network's own 10,000-visit reference. These frozen-network measurements establish the value of search depth, not an optimal training schedule.

The training cap increases through 300, 400, 500, 600, and finally 800 visits. Cheap early search supplies games quickly; deeper later search gives the improving network richer targets. Retaining the relevant subtree after a played move also saves work: some visits at the new root have already been performed during the previous move.

\subsubsection{Why fast and full searches did not transfer}\label{sec:04a-search-why-fast-and-full-searches-did-not-transfer}
A scheme inspired by KataGo's playout-cap randomization \cite{ref2} used cheap searches to advance most moves and full searches on a random minority, which alone became primary policy targets. The trade is attractive in 19\ensuremath{\times}19 Go: long games make finished outcomes expensive, so cheap moves can supply more independent value targets. Chess games are shorter, their outcomes and positions are often clearer, and the value objective was already learning. The transferred recipe thus spent search on moves whose positions were discarded as primary training rows without showing that the extra completed games compensated for the lost policy-target density.

With a quarter of moves searched to 600 visits and the rest to 150, cheap moves consumed 42.9\% of nominal search work despite being excluded as primary targets. They still selected played moves, affected terminal outcomes and retained trees, could supervise a preceding row's next-policy target, and influenced which positions entered the restart archive.

The mixed workload also interacted badly with batching. Once the cheap searches finished, only the full-search minority remained. In a 512-game test this tail left about 128 active trees and filled only 86 of 320 available batch slots with one leaf per tree. Allowing four leaves per tree raised the average batch to about 268 and improved search throughput by roughly 20\%, but it did so by making each search less serial. A policy intended to save compute thus created pressure to accept a search-quality tradeoff merely to keep the accelerator occupied.

The retained design searches every recorded move to the scheduled cap, preserving both policy targets and endgame coverage.

\subsubsection{Three attempts to allocate search adaptively}\label{sec:04a-search-three-attempts-to-allocate-search-adaptively}
The first attempt asked whether search could stop when the most-visited move looked uncatchable. An audit of completed trees found apparent room: an optimistic reconstruction suggested that a rule requiring at least 75\% of the cheap budget might remove 15.3\% of cheap-search visits, or 6.38\% of all nominal limits if full searches saved nothing. But the records contained final visit distributions, not the intermediate traces needed to identify when the leader first became safe. They also omitted enough state to reconstruct retained visits and forced-playout-pruned targets. The reconstruction could suggest where to look for savings, but it could not tell a running search when to stop.

A wider offline study found a deeper problem. Concentrated visits did not consistently mean that search was finished: very diffuse and already-decided positions gained little, while moderately concentrated but contested positions gained most. Simple concentration thresholds tended to stop in the very region where more search was useful. The rule-based approach was therefore declined before an online strength match.

The next allocator replaced the rule with a learned prediction, related in aim but not identical to dynamic simulation stopping \cite{ref3}. An auxiliary head estimated, for several candidate budgets, how far that budget's policy would differ from a deep-search policy. A correction based on the current root statistics refined this prediction, and a controller adjusted allocation to keep average spend near its target. At approximately matched mean spend it captured about 23\% of the possible reduction in policy divergence; its mean target fidelity resembled roughly 1.18 times uniform search at 0.967 times the spend. It also behaved plausibly, assigning more work to contested positions.

Learning nevertheless deteriorated. Repeated online attempts trailed comparable non-adaptive training by roughly 60--100 ladder Elo. More than a third of positions received an average budget fraction near 0.36, and almost 9\% received one eighth of baseline search. Those shallow policies were close to the network's own prior yet were trained at full weight. One explanation is that closeness to a deep policy does not measure how much a target will improve the next network. Easy-to-predict targets can agree with deeper search while teaching little beyond what the network already knows. The controller improved policy agreement, but its online tests produced worse learning.

The final adaptive system moved the decision inside search, where a learned stopper could observe the evolving tree rather than predict difficulty in advance. Its decisive test started from the same checkpoint, optimizer, and rebuilt replay state for every arm. The most aggressive setting skipped about 14\% of nominal search, and its internal credit-wait measurements changed in the expected direction. Yet generation cadence improved by only about 3\%. Self-play overlapped the optimizer, so most of the removed search was slack rather than critical-path work. The paired strength differences, calculated as baseline minus stopper, were +1.7 \ensuremath{\pm} 9.9 Elo and -4.2 \ensuremath{\pm} 10.1 Elo (standard errors) for the two settings: neither resolved a strength effect. At the observed learning rate, the cadence gain was worth only about one Elo over three hours, below the experiment's resolution.

The three approaches failed for different reasons. The threshold rule lacked an identifiable safe signal; the predicted allocator improved policy fidelity but not learning; and the in-search stopper removed mostly off-critical-path work while search and training shared the available GPU capacity.

\subsubsection{Parallel leaves: buying latency with search quality}\label{sec:04a-search-parallel-leaves-buying-latency-with-search-quality}
Batching across many independent games is the cleanest way to feed the accelerator, but the number of active roots eventually runs out. The engine can then keep several leaf traversals in flight from one root. Virtual reservations discourage those traversals from selecting the same path, yet every selection is based on a tree that is missing the other in-flight results. Parallel search is therefore not serial MCTS executed faster. It exchanges fresher decisions for larger batches and lower latency.

To measure this cost, the batch must have room for multiple leaves from each tree. Otherwise, many independent games can fill it before per-tree parallelism has any effect. A sweep that deliberately provided this room estimated a loss of 6.4 \ensuremath{\pm} 4.7 Elo per doubling of parallel leaves. With that uncertainty, comparisons at specific search budgets are more informative than treating the estimate as a general rule.

The tradeoff also depended on the total budget. Table \ref{tab:04a-search-1} pairs strength with match duration in a fixed-network sweep at 1,000 searches against 20,000-node Stockfish.

\begin{table}[!t]
\centering\small
\caption{Parallel search at 1,000 visits against 20,000-node Stockfish}\label{tab:04a-search-1}
\begin{tabular}{@{}llll@{}}
\toprule
Leaves & Elo & Time (min) & Elo change \\
\midrule
1 & 2,823 & 18.1--18.5 & 0 \\
4 & 2,804 & 3.4--3.8 & \ensuremath{-}19 \\
16 & 2,778 & 1.3--1.8 & \ensuremath{-}45 \\
\bottomrule
\end{tabular}
\end{table}
Four- and sixteen-way parallelism reduced match time by roughly factors of five and eleven. Their strength differences were smaller than the overlapping match intervals. At only 100 searches, however, sixteen-way parallelism reduced the central strength estimate by about 235 Elo. A deeper search will eventually visit more of the temporarily suboptimal leaves selected from stale state, making the same concurrency less disruptive.

The resulting choice differs between training and interactive play. Many independent self-play games can fill batches without much per-tree parallelism. A player waiting for one move has only one root, making parallel leaves more useful. In either case the budget must be large enough to tolerate the less-informed selections.

\subsubsection{When a tree became a graph}\label{sec:04a-search-when-a-tree-became-a-graph}
Chess appears rich in transpositions: different move orders often reach the same board. The project tested whether Monte Carlo graph search, as in Czech, Korus, and Kersting \cite{ref6}, could share neural evaluations, descendants, and search statistics across those paths. Canonical nodes held shared state information while parent/action edges kept local PUCT statistics. Sharing a position in this way is more demanding than looking it up in a table: results must be propagated correctly through its incoming paths, and the graph must remain usable after a played move.

The limiting fact was chess-state identity. Pieces and side to move are not enough: castling rights, en-passant state, the halfmove clock, and repetition-relevant history can change the legal result. Merging positions that differ on those fields would create an approximate algorithm with different game semantics. Under exact equality, most apparent board transpositions disappeared.

In the corrected chess self-play benchmark with a randomly initialized network, useful sharing was negligible. Only 0.0249\% and 0.1769\% of neural evaluations were avoided at 1,000 and 10,000 searches, while the graph was 8.63\% and 8.28\% slower. Structural counters confirmed that shared descendants and statistics were active. Exact reuse was too sparse to repay the bookkeeping cost in this implementation, so it was not taken to a final strength match.

\subsubsection{Why inference caching found little to reuse}\label{sec:04a-search-why-inference-caching-found-little-to-reuse}
Inference caching asked a narrower question: if an identical encoded neural input reappears under the same model, can its policy and value outputs be reused without sharing search state? Two investigations answered complementary parts of it.

First, a real bounded, sharded cache was shared by the search threads within each self-play process. With eight processes per GPU, three search threads per process, and capacity for 1.5 million entries per process, its hit rate was 0.970\%. Disabling the cache made game updates about 0.88\% faster in the short stochastic comparison and reduced summed worker peak memory by about 7.12\%. The implemented cache consumed memory without demonstrating a speedup.

A later audit measured the upper bound available to a wider cache before building one. An unbounded tracker shared across one search executor observed exact encoded inputs but deliberately evaluated every position. With diverse starts and production-style progression, repeats were 1.33\% at 150 searches, about 4.24\% at 800, and about 3.5\% in the mixed workload; same-batch duplicates were essentially absent. Following retained trees for several moves raised the rate only to roughly 4\%. The tracker itself cost about 3.65\% throughput and grew without bound. A real cache would additionally pay for output storage, synchronization, eviction, and device transfers while retaining fewer entries.

The wider design was declined before implementation. Together with the implemented per-process cache, its opportunity audit showed little reuse available under this workload.

\subsubsection{The retained search design}\label{sec:04a-search-the-retained-search-design}
Fixed visits supply a consistent amount of search to each recorded move. Exploration prevents this search from simply repeating its favourite opening: root noise perturbs the move prior, and temperature controls how strongly move selection favours the visit leader. Forced root playouts ensure that alternatives receive some attention; their visits are pruned from the training target when they reflect forced exploration rather than a genuinely preferred move.

For an unvisited move, first-play urgency starts from a reduced parent value rather than an optimistic default. A 0.99 per-ply discount favours nearer favourable outcomes, while retaining 60\% of subtree visits after a played move reuses analysis without allowing old statistics to dominate completely. Together with native batching, these mechanisms make fixed-budget search both exploratory and affordable. Appendix \ref{app:D} gives the complete numerical recipe.

\subsection{Getting more learning from each game}\label{sec:04b-data-and-replay}
\begin{figure*}[t]
\centering
\includegraphics[width=494.476bp,height=0.70\textheight,keepaspectratio]{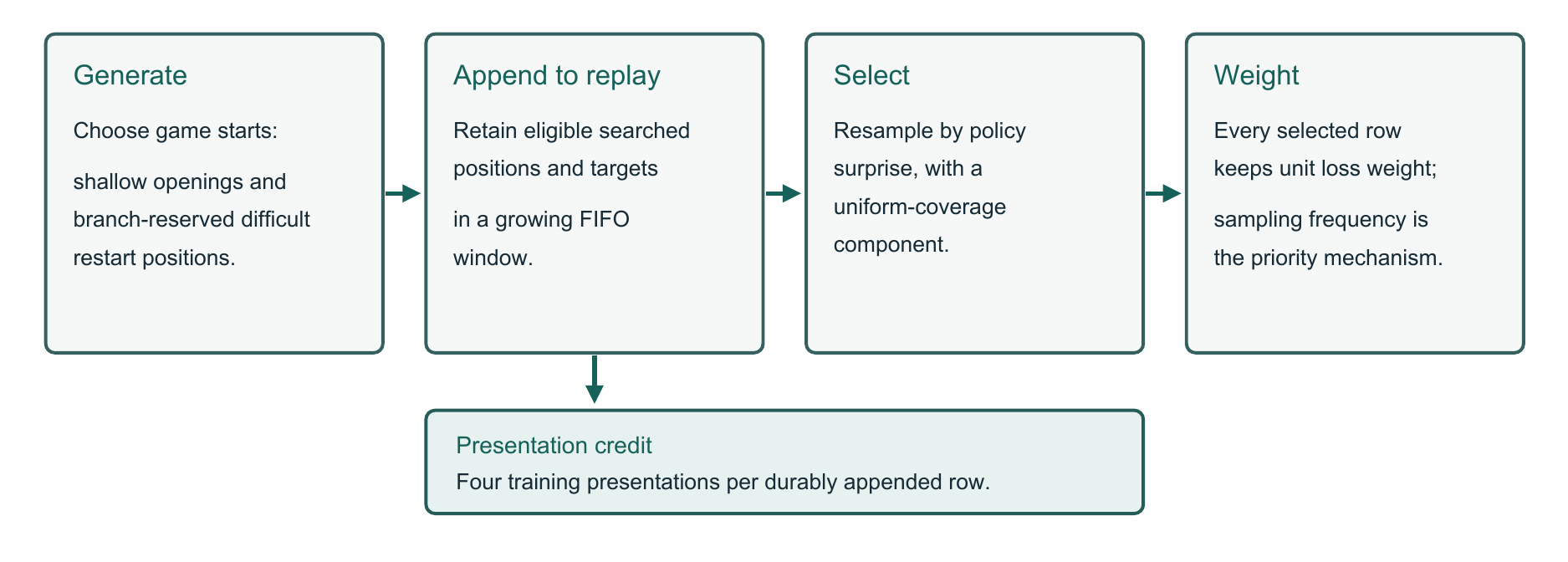}
\caption{Starting positions shape the games that are played. Their searched positions enter replay, where sampling determines which examples the learner revisits. New examples also set the pace of optimizer updates.}\label{fig:replay-decision-path}
\end{figure*}

The data strategy determines both the distribution of self-play experience and its contribution to learning. Restart states direct new games towards unresolved alternatives; replay capacity and sampling determine which searched positions remain available and how often they are trained on. Resignation and cutoff handling affect the reliability of their outcome targets. Figure \ref{fig:replay-decision-path} distinguishes these decisions along the path to a training batch.

\subsubsection{Restart-state selection}\label{sec:04b-data-and-replay-restart-state-selection}
The start distribution combines opening diversity with targeted exploration. Half of games begin after a uniformly selected zero to eight random legal plies. These prefixes diversify the opening cheaply; they are reconstructed in the recorded history but not searched or trained directly. Drawing zero plies also keeps some games at the ordinary initial position.

The other half begin from archived self-play states, falling back to random openings when a worker's restart archive is empty. The archive favors positions with a meaningful unresolved alternative: they are not near the end of the source game or already decided, and search found a small set of plausible moves. A restarted game chooses an untried branch, reconstructs the prefix, then explores the alternative the source game did not play. Appendix \ref{app:D} gives the eligibility thresholds.

The archive gives 30\% probability to uniform selection. Otherwise it favors positions where search substantially corrected the network's value estimate, while softening the priority so one extreme position cannot dominate. Age and capacity bounds remove old states, and exhausted positions leave the archive. Archives are local to workers.

Starting from archived states follows the search-control idea studied by Trudeau and Bowling \cite{ref5}. Here, observed search disagreement and untried branches guide the archive. Playing those branches creates new trajectories; the replay sampler below instead chooses which existing positions recur during training.

\subsubsection{Replay admission}\label{sec:04b-data-and-replay-replay-admission}
Replay admits searched positions, including a final search used to evaluate a capped game's endpoint. Random opening moves and reconstructed restart prefixes have no search target and are excluded. Admission does not depend on ply: searched endgame positions remain eligible throughout the game. Sparse policy storage retains at most 60 actions per position and records the discarded visit mass.

\begin{figure*}[t]
\centering
\includegraphics[width=486.423bp,height=0.70\textheight,keepaspectratio]{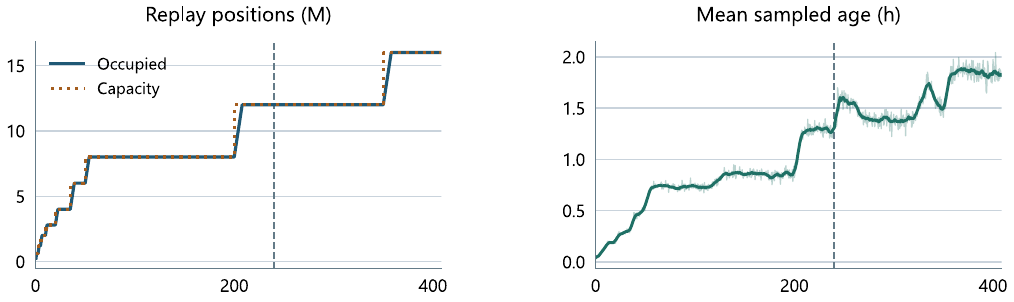}
\caption{Replay occupancy follows the expanding capacity during final training. The mean age of sampled positions grows to roughly two hours; age is measured from completion of the source game. Both horizontal axes show completed optimizer steps in thousands. The dashed line marks the small-to-medium model transition.}\label{fig:appendix-replay-age}
\end{figure*}
\subsubsection{Replay capacity and reuse}\label{sec:04b-data-and-replay-replay-capacity-and-reuse}
The capacity experiments addressed the diversity--age tradeoff introduced in Section \ref{sec:03-system-and-methods-replay-and-materialization}.

The physical memory map is preallocated once, while its logical capacity grows from 600,000 to 20 million rows. The early window stays small when little data exists; later growth preserves more openings, endgames, and policy history. In an earlier campaign, playing strength continued to improve after fixed-dataset policy accuracy largely saturated, reinforcing the value of fresh and varied positions. Appendix \ref{app:D} gives the full capacity schedule. Figure \ref{fig:appendix-replay-age} shows replay occupancy and sampled-position age during final training.

Replay reuse introduces a second tradeoff. The configured ratio is the number of optimizer presentations funded by each newly admitted row. In the retained setting, four presentations are credited per row; a 500-step quantum at a global batch of 2,048 therefore requires 256,000 newly appended rows. Only positions that have actually reached the replay store count towards this allowance.

Higher reuse funds more updates from each game; lower reuse gives each update fresher positions if the actors can supply them. Short controls at ratios 4, 6.25, and 8 showed similar strength despite different update rates. The selected ratio of 4 favors freshness. Appendix \ref{app:C} gives the duration and scope of these short controls.

\subsubsection{Policy-surprise sampling}\label{sec:04b-data-and-replay-policy-surprise-sampling}
The retained sampler assigns seventy percent of draws to policy surprise, capped at 2.0 so extreme rows cannot dominate; the remaining 30\% are uniform to preserve coverage. A row can recur across optimizer steps but is drawn only once within a global batch. The closer precedent is KataGo's policy-surprise weighting \cite{ref7}, which increases sample frequency according to disagreement between the search target and policy prior. This also belongs to the broader family of prioritized replay \cite{ref4}.

The priority deliberately changes which positions dominate training. It is not corrected back to uniform sampling. Because surprise can also reflect search noise or old targets, the uniform component preserves broader coverage.

Priority acts through sampling frequency, not through per-example loss weights: every sampled row has weight 1.0. An earlier replay design instead weighted merged duplicate positions by their multiplicity.

\begin{figure*}[t]
\centering
\includegraphics[width=494.476bp,height=0.70\textheight,keepaspectratio]{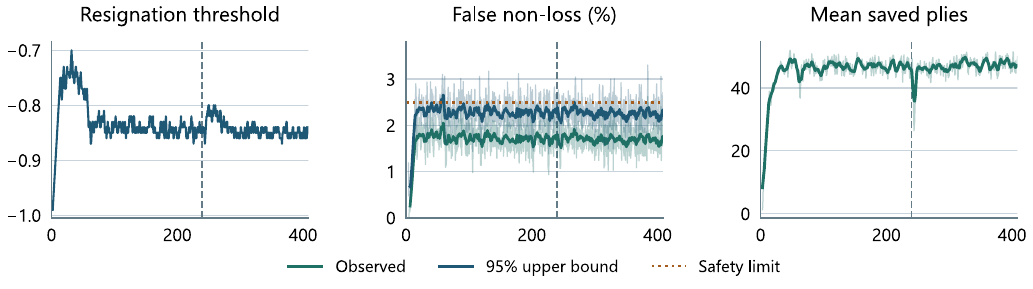}
\caption{Resignation threshold, observed false-resignation rate and its one-sided 95\% upper bound, and estimated saved plies during final training. The error panel shows periods when the resignation gate is enabled; the dotted line marks the 2.5\% calibration target, including observed excursions. Saved plies are estimated from continuation games. Faint traces show raw values, with eleven-quantum averages overlaid. All horizontal axes show completed optimizer steps in thousands; the dashed vertical line marks the small-to-medium model transition.}\label{fig:appendix-resignation}
\end{figure*}
\subsubsection{Resignation calibration}\label{sec:04b-data-and-replay-resignation-calibration}
Resignation can save a large amount of search, but a false resignation turns a drawable or winning position into an incorrect terminal label. A fixed value threshold was therefore replaced by continuing measurement. Twenty percent of games are designated at creation as continuation games and can never resign. They reveal the outcomes of hypothetical triggers for thresholds from -0.99 through -0.70. A trigger requires both the root value and the best visited child's backed-up value to cross the threshold; evidence from capped games is excluded because their natural outcome remains unknown.

Calibration estimates the fraction of hypothetical resignations whose continued games end in a draw or win. A threshold requires at least 100 observations and a one-sided 95\% upper confidence bound no greater than 2.5\%. It can become more conservative immediately but relaxes by at most 0.01 per publication. Recalibration tracks changes in the model's value predictions. Figure \ref{fig:appendix-resignation} reports the threshold, rolling error estimates, and saved plies.

\subsubsection{Outcome targets at the ply cap}\label{sec:04b-data-and-replay-outcome-targets-at-the-ply-cap}
A ply cap creates a harder target problem because there is no observed result at all. The project compared material heuristics, raw network value, values from adjacent searches, and a fresh search at the cut position. Playing games beyond the normal cap supplied later outcomes for evaluation. Brier score measures squared probability error, while cross-entropy penalizes assigning low probability to the eventual outcome; lower is better for both. At the earlier measured checkpoint, the cut-position search achieved Brier score 0.444 and cross-entropy 0.756, compared with 0.491 and 0.851 for material divided by 39. At the later checkpoint the corresponding scores were 0.193 and 0.374 versus 0.388 and 0.707. Among decisive continuations, search-root sign accuracy exceeded 98\% in both cohorts; calibration, not merely sign, distinguished the targets. This comparison motivated replacing material-based targets with searched values. Initially, the worker reused the preceding move's root value, which could come from a forced cheap search. The retained worker instead performs a dedicated full search at the actual cut position and uses its root value as the bootstrap. The root value becomes a soft win/draw/loss target, then materialization applies the configured per-ply blur. Appendix \ref{app:D} gives the conversion.

\subsubsection{Discounting and intermediate value targets}\label{sec:04b-data-and-replay-discounting-and-intermediate-value-targets}
Discounting was introduced to favour faster conversion when many games approached the 250-ply cap. Search multiplies backed-up values by 0.99 per ply, reducing the value of a more distant win relative to an otherwise equivalent nearer win. Training applies a separate factor of 0.998 per remaining ply by blending the outcome's WDL target towards uniform. This attenuates the expected value of distant outcomes and also reduces the certainty assigned to early positions. With signed values, discounting likewise makes a delayed loss less negative.

The two discounts act at different points in the learning loop, but share the intended incentive to complete winning games sooner. Their independent benefit was not established. Restoring searched endgame positions removed a known gap in training coverage. Conversion recovered after several concurrent changes, so their individual contributions were not isolated (Section \ref{sec:05a-three-failures-late-game-target-poisoning}). Whether either discount improves the retained recipe remains open.

A separate scheduled blend incorporates up to 10\% of the stored search-root value into the training target, combining the completed game's outcome with search's assessment of the current position.

\subsubsection{Auxiliary trajectory targets}\label{sec:04b-data-and-replay-auxiliary-trajectory-targets}
For the auxiliary objectives introduced in Section \ref{sec:03-system-and-methods}, label availability depends on the recorded trajectory. A next-policy label exists only if there is a following searched observation, and a capped game cannot reveal its true remaining length. Missing labels are excluded from the corresponding loss.

Broader auxiliary bundles were set aside while diagnosing several simultaneous training problems, not because a controlled ablation found them harmful. Future-derived labels must come from complete trajectories and be masked whenever the future is unknown.

\subsubsection{Reanalysis and publication cadence}\label{sec:04b-data-and-replay-reanalysis-and-publication-cadence}
Reanalysis refreshes stored policy targets by searching their positions with the latest network. It competes for compute with fresh self-play, which supplies new positions as well as outcomes. A bounded synchronous implementation was integrated into an earlier replay design but was not retained when that design was replaced. Its learning return relative to fresh self-play was not measured.

Publishing the latest model more often can also make targets fresher, because self-play then uses more recent predictions. Doing so every 100 optimizer steps spent too much time saving, exporting, checking, and activating models. The retained 500-step interval reduces that overhead while still refreshing the actors regularly.

\subsubsection{Actor-trainer overlap}\label{sec:04b-data-and-replay-actor-trainer-overlap}
The retained schedule keeps half the actors active during each 500-step training block. Chapter \ref{sec:05-systems-optimization} presents the overlap comparison and its effect on training duration and concurrent search throughput.

\subsubsection{The retained data strategy}\label{sec:04b-data-and-replay-the-retained-data-strategy}
The retained strategy combines broad opening coverage with targeted restarts, an expanding replay window, and policy-surprise sampling. Fresh-game supply limits reuse, while calibrated resignation and searched cutoff values reduce the cost of completing trajectories. These mechanisms control the distribution, age, and reliability of the training data; their combined effect, rather than replay volume alone, determines its value to the learner.

\subsection{Choosing what the network predicts}\label{sec:04c-networks-and-training}
\begin{figure*}[t]
\centering
\includegraphics[width=494.476bp,height=0.70\textheight,keepaspectratio]{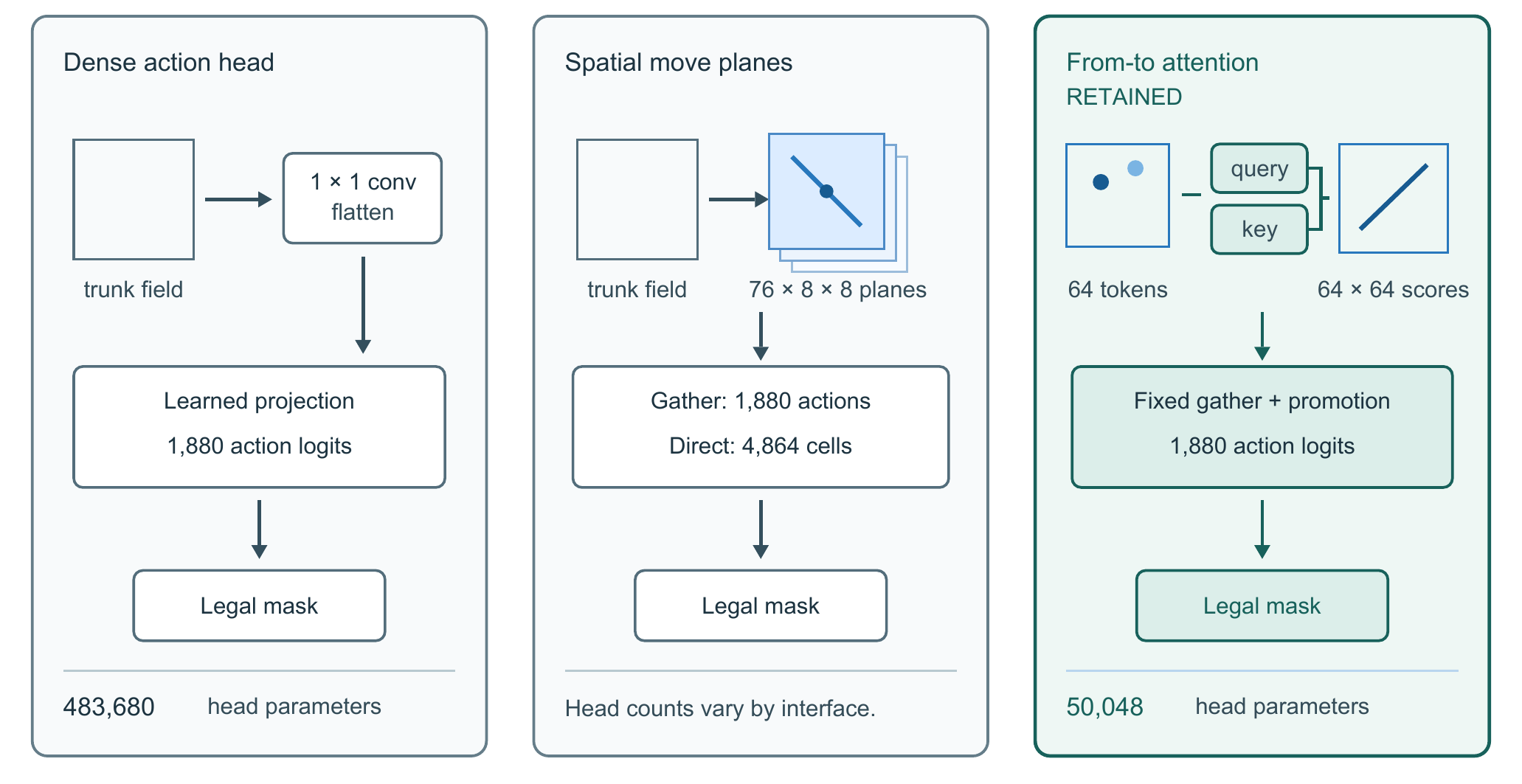}
\caption{Dense heads project to a move list, move-plane heads predict spatial move types, and the retained from-to head scores square pairs before gathering canonical actions and masking illegal moves.}\label{fig:policy-representations}
\end{figure*}
Network architecture constrains both prediction quality and the volume of search affordable during training. We compared policy representations, convolutional and attention trunks, global context, and value heads, then examined model growth and quantization. The policy head was a major source of parameter cost in small networks; Figure \ref{fig:policy-representations} summarizes the three representations evaluated.

\subsubsection{Three policy representations}\label{sec:04c-networks-and-training-three-policy-representations}
All three policy families were implemented against the chess move interface, whose selected encoding contains 1,880 canonical actions. Controlled fitting experiments compare held-out policy cross-entropy in nats.

The \textbf{dense head} flattened a small spatial projection into 1,880 logits. Dense heads trained successful models, but on a small trunk their projection could dominate the parameter count. Variants changed projection width, spatial reduction, and final-map rank. A rank-96 variant reduced head size from roughly 484,000 to 207,000 parameters.

The \textbf{move-plane head} represented sliding directions, knight moves, and promotions as spatial move types. It could preserve board structure, but its experiments did not show an advantage over the dense control. In a short supervised comparison, the repaired plane head reached 2.1828 held-out policy cross-entropy after about 2,500 steps versus 2.0824 for the dense control, while its curve was still improving faster. An online plane-head trial was retired after slower learning and weaker play.

The two implementations differed: one gathered 1,880 canonical logits from 76 planes (56 sliding directions, eight knight moves, and 12 promotions); the other exposed all 4,864 plane-square cells and normalized only legal moves. Both express the same basic idea: predict move types at board locations rather than score a flat action list.

The retained \textbf{from-to head} uses query-key dot products over the 64 trunk squares, with a fixed gather into the canonical action space and separate promotion offsets. This retains spatial structure without a large dense projection. On the controlled 128-channel convolutional trunk, the head used 50,048 parameters rather than 483,680 for the dense alternative. Section \ref{sec:03-system-and-methods-chess-representation-and-outputs} and Appendix \ref{app:D} specify its representation and action mapping.

Holding that trunk fixed, the from-to head improved the held-out policy gap by 0.0298 nats (paired 95\% interval 0.0285--0.0311 over held-out positions, not training seeds). Spending the saved parameters on a wider trunk added only 0.0018 nats in the three measured cells; the missing fourth cell prevents fully separating the two effects. Forward throughput fell by approximately 1.9\% at batch 512 and 9\% at batch 64. The from-to head therefore improved policy fit and reduced parameter count, with a serving penalty that was smaller at the large batches used for self-play.

\subsubsection{Board input and rule state}\label{sec:04c-networks-and-training-board-input-and-rule-state}
The selected model receives 52 side-to-move-canonical board planes: pieces, castling rights, en passant, checks, repetition, the eight most recent moves, material counts, and the fifty-move counter. Rule-sensitive planes keep positions with different legal or draw states distinguishable. Appendix \ref{app:D}, Table \ref{tab:appendix-d-reproducibility-1} lists all 52 planes and their encodings. File reflection is the only augmentation; it also mirrors action targets and exchanges kingside and queenside castling planes. Some earlier component comparisons used a 29-plane input, so their absolute scores are not input-matched to the final model.

\subsubsection{Convolution, attention, and global context}\label{sec:04c-networks-and-training-convolution-attention-and-global-context}
The trunk comparison evaluated whether attention improved board-wide feature learning enough to justify its serving cost. Both designs shared their trunk across policy, value, and auxiliary heads. Because dense primary and auxiliary policy heads could dominate a small model's parameters, head choice had to be controlled alongside the trunk.

The attention alternative treated the 64 squares as tokens, with learned row and column embeddings, pre-normalized self-attention, and GELU feed-forward blocks. No-bias, relative-offset, and input-dependent Smolgen-style attention biases were implemented. These biases respectively add no positional preference, depend on the offset between squares, or depend on the board itself. The preserved comparison covers the first and third.

With bootstrap policy shape, head, runtime, and precision controlled, the convolutional trunk beat bare attention by 0.0060 nats on held-out teacher data. Replacing the attention model's dense head with from-to and reallocating capacity to the trunk improved the held-out gap by 0.1573 nats; this was not an isolated head replacement. The best Smolgen attention cell achieved slightly better held-out fit than the convolutional comparison, by 0.0090 nats. Convolution nevertheless remained the preferred operating choice because attention served more slowly and consumed more memory at the relevant batch sizes. Appendix \ref{app:C} gives the serving comparison.

Local convolution receives global context every second residual block: board-wide means and maxima from one quarter of the channels are projected back as biases on local features. Squeeze-excitation was another implemented context mechanism, using board-wide information to rescale channels. Global pooling appeared to learn faster, with no clear final-strength difference; this remains a qualitative observation because the comparison results are unavailable. The pooled features give local convolutions access to the whole board, following the KataGo precedent \cite{ref2}. Those context operations remain in floating point when the main convolutions are quantized.

\subsubsection{Value and training-only heads}\label{sec:04c-networks-and-training-value-and-training-only-heads}
The outcome head predicts win, draw, and loss rather than a single scalar. Search converts this distribution to an expected value when necessary, while training and diagnostics retain the draw probability. A compact two-channel spatial reduction and 48-unit hidden layer was retained. A matched 32-channel probe added 97,020 parameters and reduced measured training throughput by 1.31\%, while improving total loss by only 0.00309 in one short seed and slightly worsening WDL loss. This justified keeping the smaller head. The earlier scalar-to-WDL transition has no preserved isolated strength comparison.

Training also predicts the next move's searched policy and normalized remaining game length, with loss weights 0.15 and 0.1. The next-policy target uses the following state's own side-to-move action space, legality mask, and symmetry transformation. Both heads train the shared representation without making inference more expensive, because deployment removes them. Other auxiliaries were set aside while debugging potential interference, not because they were shown harmful. The retained pair has not received a long matched self-play ablation.

\subsubsection{Bootstrap and optimization controls}\label{sec:04c-networks-and-training-bootstrap-and-optimization-controls}
Initialization shapes the first search priors and, through them, the first self-play targets. In an initial architecture comparison, the attention policy put only about 0.11 probability mass on its top three moves while the convolutional control was effectively one-hot. Neither extreme represented chess knowledge, but each induced different data. The retained bootstrap uses architecture-appropriate initialization, deterministic construction, a small final policy projection, and calibration on 516 encoded positions toward a common policy concentration. Those positions set numerical scale without supplying supervised chess targets.

The retained optimizer is Nesterov SGD. Frozen-replay tests helped choose warm-up and learning-rate settings before committing to self-play runs. AdamW had also trained successful models; the choice of SGD does not mean AdamW failed. Gradient clipping limits unusually large updates. The final run completed 408,500 optimizer steps, with the learning rate, losses, and gradient norms shown in Appendix \ref{app:A}.

\subsubsection{Quantization as an architectural constraint}\label{sec:04c-networks-and-training-quantization-as-an-architectural-constraint}
INT8 deployment required changes to the residual architecture as well as quantization-aware training (QAT). The experiments compared prediction fidelity and compiled throughput: an architecture that tolerates rounding but introduces expensive precision conversions may still be unsuitable for self-play.

Post-training quantization of the ordinary residual tower accelerated inference but severely distorted predictions. Activation ranges grew from about 0.78 near the input to roughly 40--43 late in the network; depending on calibration, full-trunk INT8 preserved only 13.3--25.9\% policy top-one agreement. Weight-only quantization was faithful but slower than TensorRT FP16. Quantization therefore became a network-design problem rather than a final export switch.

A scaled pre-activation block bounded residual activation ranges, but its normalization, clipping, scaling, and repeated precision conversions fragmented the compiled TensorRT graph. It expanded an 83-layer FP16 graph to 330 layers and an INT8 graph to 470 layers, with many reformats, yet still failed fidelity. The retained scaled post-activation block instead keeps the conventional sequence of convolution, normalization, capped activation, convolution, normalization, scaled residual addition, and capped activation. Activations are capped at six, and residual branches are scaled by the inverse square root of depth. The scale can be folded into the second convolution at export, which preserves more efficient compiler tactics.

Batch-normalization folding changed the quantization behaviour as well as the compiled graph. The final run keeps the trainable model unfused and performs folding and recalibration on a deployment copy. Separate fitting experiments also evaluated continued training in the folded topology.

The scaled post-activation block learned normally under quantization-aware training. After continuation in the folded deployment topology, a production-sized smoke test reached roughly 135,000 INT8 positions per second, compared with 60,000 for TorchScript BF16 and 99,000 for TensorRT FP16. These are model-core rates, not end-to-end self-play rates. Folding only after training damaged agreement, and global context and the heads remain outside the INT8 trunk. The result is evidence for an architecture compatible with efficient INT8 inference, rather than a playing-strength advantage over the ordinary floating-point residual block.

\subsubsection{Progressive model sizing}\label{sec:04c-networks-and-training-progressive-model-sizing}
\begin{figure*}[t]
\centering
\includegraphics[width=494.476bp,height=0.70\textheight,keepaspectratio]{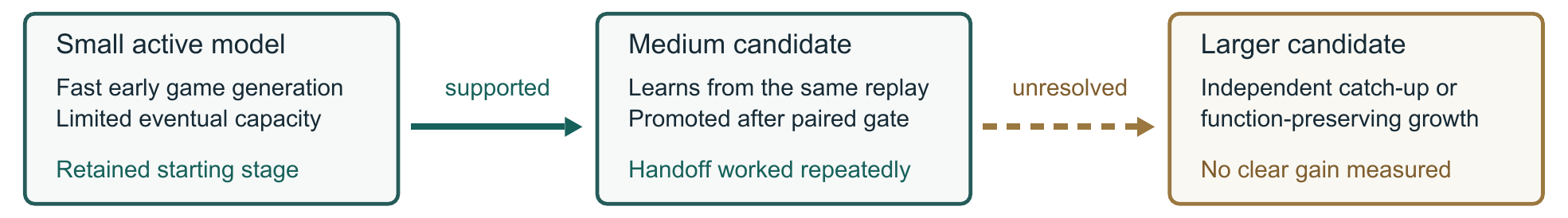}
\caption{A small model reduces early self-play cost, and a medium candidate trains on the same replay before paired-match promotion. The larger candidate may avoid catch-up with function-preserving growth, but the limited continuation did not demonstrate a strength gain; the reported checkpoint remains medium-sized.}\label{fig:progressive-model-sizing}
\end{figure*}
Following KataGo \cite{ref2}, progressive sizing uses a small network for early self-play while a larger candidate trains on the same replay before promotion. The throughput advantage is most useful while the smaller network has enough capacity to absorb the available experience. The small-to-medium transition worked repeatedly in this project.

Candidate start follows a stage-specific searched-Elo plateau; promotion instead requires two passing paired matches against the active model. Section \ref{sec:05a-three-failures-promotion-from-incomparable-training-losses} explains why matches replaced training loss as the promotion criterion; Appendix \ref{app:D} states the retained thresholds.

An independently initialized larger candidate needed substantial catch-up. Function-preserving growth instead initializes the larger model to compute the same predictions as the medium model. It avoids relearning that function, but might also bias optimization toward the smaller model's existing representation. The limited grown-model continuation reached parity without a clear strength gain. The reported model therefore remains medium-sized; choosing between catch-up and growth needs a matched-compute comparison.

\subsubsection{Distillation and compact models}\label{sec:04c-networks-and-training-distillation-and-compact-models}
Distillation evaluated the tradeoff between reduced inference cost and approximation of a stronger teacher. We tested two sources of supervision: the larger teacher's direct predictions and the targets already collected in replay. In the first, students learned the teacher's legal-move probabilities and WDL predictions from games played without search. Increasing this dataset from one million to six million positions mattered more than a small capacity sweep. The strongest 1.33-million-parameter student trailed its teacher by 176.1 Elo at 25 searches each and by 38.4 Elo when it received the measured shallow equal-compute allowance of 58 searches. At 250 searches each the gap widened to 257.6 Elo. Deeper search amplified the better prior rather than washing out approximation error. Fixed-ply sampling gave the dataset only one side-to-move parity, and the teacher had a known conversion weakness, so these absolute gaps do not transfer to the final model.

Replay-target compression instead trained on sparse MCTS visits, outcomes, root values, and replay metadata from a frozen ten-million-row window. The selected 474,069-parameter student was 13.20 times smaller than its teacher. It trailed by 291.3 Elo at 64 searches each and by 166.2 Elo when its measured saturated serving advantage allowed 186 searches against 64. It reached statistical parity only at an equal-multiply-accumulate allowance of 850 searches, which counted neural arithmetic but omitted tree work and inference overhead. The smaller model's theoretical compute advantage was therefore much larger than the extra search it could actually perform in the same time.

A final compression study trained a 470,295-parameter student on a separate frozen 20-million-row replay snapshot for 110,000 optimizer steps, roughly 23 epochs. Chapter \ref{sec:07-final-run-results} presents these final student results alongside the full model.

The resulting compact models reduced inference cost, but their realized search advantage was substantially smaller than their parameter or arithmetic ratios. The teacher's strength advantage also increased with search depth in the measured comparisons. Distillation was separate from the primary self-play training run.

\subsubsection{Decision}\label{sec:04c-networks-and-training-decision}
The retained network combines the rule-aware 52-plane input, a shared convolutional trunk, periodic global context, the from-to policy head, a compact WDL head, and training-only next-policy and remaining-length objectives. Scaled post-activation blocks make the trunk compatible with quantization-aware deployment. Progressive sizing successfully exploited a small model before handing off to the medium model, while the value of the larger stage remains unresolved. Component tests support parts of this design, while final playing strength belongs to the assembled system.

\section{From inference speed to learning speed}\label{sec:05-systems-optimization}
\begin{figure*}[t]
\centering
\includegraphics[width=494.476bp,height=0.70\textheight,keepaspectratio]{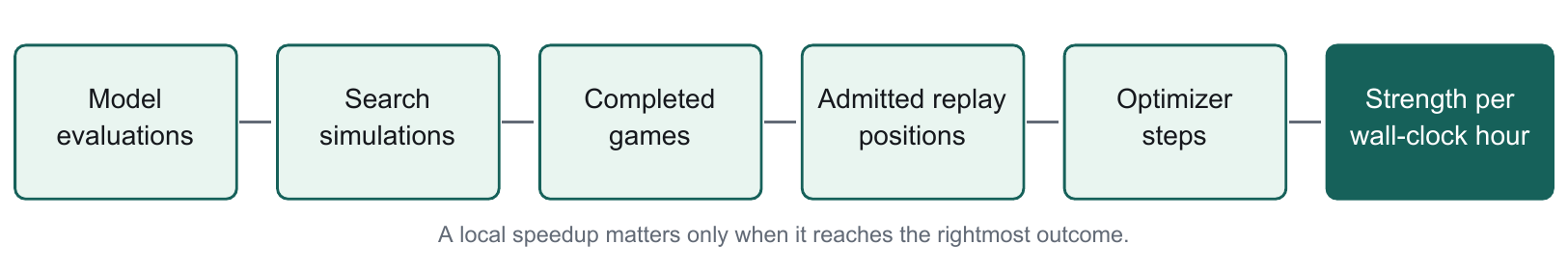}
\caption{Search speed passes through game completion, replay admission, and optimizer work before it can affect playing strength. Each boundary has its own throughput measure.}\label{fig:throughput-to-learning}
\end{figure*}

Under a fixed compute budget, inference throughput matters through the training data it makes affordable. Search, game completion, replay delivery, and optimization jointly determine that supply (Figure \ref{fig:throughput-to-learning}). This chapter examines the bottlenecks at those stages and the scheduling required to share GPUs between actors and the learner. Relative gains are reported within each controlled benchmark; Appendix \ref{app:C} collects the absolute rates and settings.

\subsection{Native ownership of the search loop}\label{sec:05-systems-optimization-native-ownership-of-the-search-loop}
Per-position Python dispatch made board updates, legal-move generation, encoding, and inference submission a host-side bottleneck. The native execution boundary described in Section \ref{sec:03-system-and-methods-why-the-search-loop-stays-in-c} removes that per-position dispatch.

Each actor interleaves 512 games through a preallocated inference pipeline and retains search subtrees across moves. The retained topology uses four actors per GPU, one inference worker per actor, batches of up to 320 positions, and two outstanding batches. This provides independent work to overlap tree traversal with neural evaluation without introducing a Python boundary at each leaf.

\subsection{Batching and host-side submission}\label{sec:05-systems-optimization-batching-and-host-side-submission}
Batching amortizes GPU launch costs, but sustained utilization also requires timely host-side preparation. Preallocated staging buffers, asynchronous copies, and completion events overlap encoding and submission with GPU execution. CUDA graph replay reduces the remaining launch overhead.

In a controlled 32-process self-play workload, the optimized path increased search throughput by 20.5\% while reducing aggregate actor CPU consumption from 52.6 to 19.8 cores. Average inference batch size rose from 141 to 222 and the number of model calls fell by 24\%. On the same node restricted to 24 CPU cores, throughput more than doubled because reducing submission overhead allowed the previously starved GPUs to remain busy. The larger gain under the CPU quota identifies host submission as a substantial constraint on GPU utilization.

Full batches were necessary but insufficient for saturation. Reducing TensorRT actors from four to two per GPU lowered search throughput by 39\% despite full individual batches, because fewer actors reduced overlap between CPU preparation and GPU inference. Additional inference threads instead duplicated CUDA contexts and fragmented batches. The measurements supported process-level concurrency with one inference thread per actor.

Per-tree parallelism can supplement batching when too few independent games remain active. Unlike inter-game concurrency, it changes leaf selection through virtual reservations and can reduce search quality. Its strength-throughput tradeoff is examined in Chapter \ref{sec:04-research-investigations}.

\subsection{Inference runtimes and precision}\label{sec:05-systems-optimization-inference-runtimes-and-precision}
With host submission sustained, runtime and precision determine the cost of evaluating each batch. A TensorRT FP16 engine delivered 1.86x the inference throughput of a TorchScript BF16 control in a matched benchmark. Quantization-aware INT8 added a further 1.31x over TensorRT FP16 on the tested quantization-oriented network. Production-topology tests found INT8 gains of 14.4\% for the smaller network and 39.1\% for the medium network.

As Section \ref{sec:04c-networks-and-training} explains, the INT8 speedup required training the network to tolerate quantization. Most trunk convolutions run in INT8, while the start block, heads, and linear layers remain at higher precision. Export records the quantization and dequantization operations explicitly in ONNX so TensorRT can compile the intended arithmetic. Rather than rebuild the complete engine after every update, publication refits a prepared template with the new weights \cite{ref8}. Chapter \ref{sec:05a-three-failures} examines a failure in this step that made output comparisons essential.

The alternative \texttt{torch.compile} path accelerated eager batch-64 inference by roughly 27--33\%, but fused TorchScript remained faster. In the tested eight-GPU training workload, compilation reduced throughput by about 18\% relative to eager execution, while bfloat16 autocast improved it by 9.2\%. Compilation was not retained for production inference or training on this workload.

Similar parameter counts did not imply similar inference cost. Width and depth changed kernel efficiency, TensorRT tactics, and memory behavior discontinuously. Channels-last layout and cuDNN autotuning helped relevant CNN shapes, but no analytic parameter-count rule predicted the fastest network. Progressive model sizes were therefore benchmarked at their actual serving batch and precision rather than selected from FLOPs alone. The width and depth sweeps in Appendix \ref{app:C}, Table \ref{tab:appendix-c-supporting-comparisons-1} and Table \ref{tab:appendix-c-supporting-comparisons-2}, quantify this mismatch: narrower networks were not consistently faster, and changing batch size could reverse the ranking of deep-narrow and shallow-wide designs.

\subsection{Replay materialization and training supply}\label{sec:05-systems-optimization-replay-materialization-and-training-supply}
Replay delivery must sustain both trajectory ingestion and training-batch retrieval. Parallel materializers convert completed trajectories into a circular memory-mapped store, avoiding repeated trajectory decoding in the training path. Direct column access and pinned-memory prefetching reduce batch preparation and transfer costs.

A loader benchmark on 2.5 million rows became 8.26x faster after compacting 5,000 small producer shards into 25 containers, putting delivery capacity 44\% above the measured trainer demand. In a live interval, materialization could append positions more than six times as fast as self-play supplied them. The replay pipeline therefore had enough headroom to keep up with game production.

The configured reuse ratio couples optimizer progress to newly admitted positions, so loader capacity beyond trainer demand does not by itself increase training volume. Persistent distributed trainer processes avoid startup and model construction costs between blocks. The retained global batch of 2,048 uses bfloat16 autocast across eight GPUs. Larger batches improved hardware throughput in the benchmark but also changed the number of optimizer updates per training position, so batch size was selected as part of the learning recipe rather than for throughput alone.

\subsection{Overlapping self-play and training}\label{sec:05-systems-optimization-overlapping-self-play-and-training}
Self-play and training compete for GPU capacity but have different resource profiles. Search includes CPU traversal and transfer intervals that allow useful overlap with optimizer work. Scheduling must therefore balance the slower training block against the reduction in subsequent waiting for new games.

The overlap sweep compared keeping 8, 16, or all 32 actors active during training. Moving from half to all actors active nearly doubled the trainer's work time for only a small reduction in the complete cycle. The retained half-active policy balances ongoing game production against optimizer throughput; Table \ref{tab:05-systems-optimization-1} gives both sides of this tradeoff. Cycle times are estimated from measured training duration and search throughput. With all actors paused, training alone reached 25.3 thousand samples/s.

\begin{table}[!t]
\centering\small
\caption{Actor overlap: optimizer throughput and concurrent search}\label{tab:05-systems-optimization-1}
\begin{tabular}{@{}llll@{}}
\toprule
Actors & Train (k/s) & Search (k/s) & \shortstack[l]{Estimated\\cycle time (s)} \\
\midrule
8 & 21.5 & 506 & 117 \\
16 & 17.1 & 606 & 113 \\
32 & 9.21 & 742 & 111 \\
\bottomrule
\end{tabular}
\end{table}
The useful overlap fraction depends on search cost. As visit budgets and model size rise, self-play becomes more expensive relative to training; actor settings measured at the beginning cannot simply be extrapolated to later stages.

\subsection{Throughput allocation within the learning loop}\label{sec:05-systems-optimization-throughput-allocation-within-the-learning-loop}
The resulting system allocates throughput across data generation and optimization rather than maximizing either in isolation. Native search and batched inference increase game supply, replay materialization prevents delivery from limiting training, and actor overlap replenishes replay during optimizer updates.

Additional search capacity can support more games at a fixed budget or deeper searches per position. Those uses alter data diversity and target quality differently, while replay reuse controls their rate of consumption. The systems improvements therefore expand the feasible training regime; the recipe determines how that capacity is spent. Chapter \ref{sec:06-final-chess-recipe} evaluates the resulting progress in playing strength over wall-clock time.

\section{Three failures that changed the method}\label{sec:05a-three-failures}
Three failures exposed dependencies that local optimization metrics did not capture: endgame sampling altered the reliability of value targets, TensorRT refitting changed the deployed policy, and loss-based promotion confused replay fit with playing strength. Their causes and corrections illustrate why data generation, deployment, and model selection must be evaluated as parts of the learning process.

\subsection{Late-game target poisoning}\label{sec:05a-three-failures-late-game-target-poisoning}
To prevent noisy positions from exceptionally long endings from dominating replay, an earlier self-play design excluded the late-game tail from primary training targets and reduced its search budget. This saved compute on positions that would otherwise be discarded, but systematically removed examples of endgame conversion. With little training on those positions and only shallow search during play, the model frequently failed to convert advantages before the ply cap. In the affected run, the preceding cheap search's root value supplied the outcome target for the recorded trajectory, after reversing the side-to-move perspective.

\begin{figure*}[t]
\centering
\includegraphics[width=494.476bp,height=0.70\textheight,keepaspectratio]{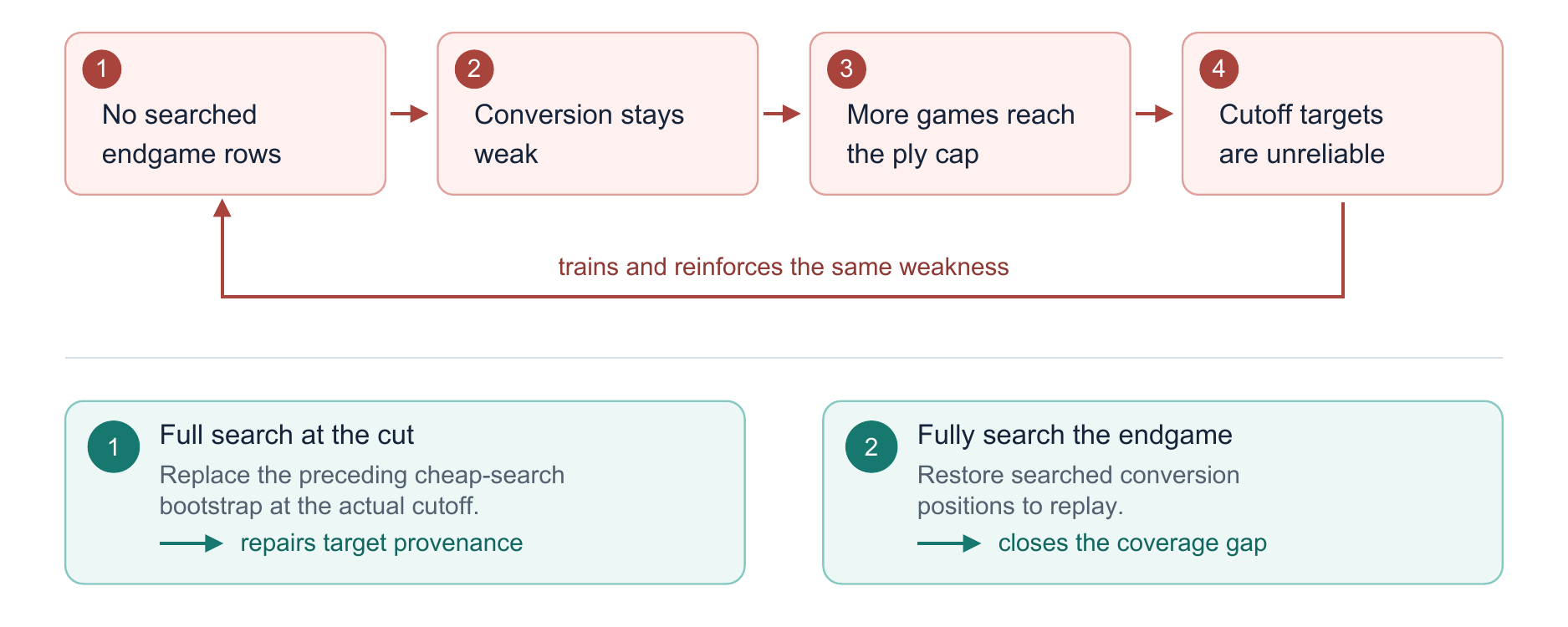}
\caption{A searched cutoff value improves labels for unfinished games; restoring fully searched endgames also returns the examples needed to learn conversion.}\label{fig:late-game-poisoning-feedback-loop}
\end{figure*}
The interaction was self-reinforcing: inadequate conversion produced more capped games, whose unreliable targets further degraded the policy and value estimates needed to finish them. During the most affected interval, approximately one ply in seven was ineligible for training, 27--32\% of games reached the cap, and 36--38\% of admitted rows inherited a cutoff target. Consequently, an error introduced at the end of a game affected training throughout its recorded trajectory. Replay turnover removed the original rows but did not necessarily remove their influence on the model or the self-play distribution.

An earlier target-comparison study had already favoured searched values over the material baseline: on 2,282 early-cut positions, the cut-position search achieved Brier error 0.444 and cross-entropy 0.756, versus 0.491 and 0.851 for material. A later set of 1,144 positions showed the same ordering. These measurements compare target estimators, not the complete conversion repair. The initial implementation reused the preceding move's search; under forced late-game cheap search, that bootstrap still came from a shallow search at the adjacent position.

The subsequent correction performed a dedicated full search at the actual cutoff, supplying both its root value and a searched policy target for that position. Restoring searched endgame positions removed a known gap in training coverage (Figure \ref{fig:late-game-poisoning-feedback-loop}). Conversion recovered after several concurrent changes, including discounting, ply-cap scheduling, and resignation settings, so their individual contributions were not isolated.

\subsection{Prediction drift under TensorRT refitting}\label{sec:05a-three-failures-prediction-drift-under-tensorrt-refitting}
The deployment pipeline refits a compiled TensorRT template after each training block to avoid rebuilding the engine. In TensorRT 10.14.1.48, a failing template's equal quantization-scale constants allowed optimizations that became invalid when subsequent training produced unequal scales. TensorRT accepted all replacement weights and reported a successful refit, yet the resulting engine no longer reproduced the source network's predictions.

On 516 real positions, the faulty engine's legal-policy KL divergence from the source network was approximately 1.05, and repeated refits with identical inputs changed individual logits by 10--15. Lowering the optimization level or making the template's scale constants distinct before compilation reduced KL divergence to approximately 0.0012 and restored deterministic refitting. The defect required the combination of equal source scales, optimization level four or higher, and a subsequent refit with unequal scales.

Template construction now separates equal quantization scales before optimization and uses a lower default optimization level. Deployment validation compares top-one legal-move agreement, policy KL, and WDL error against the source network. These checks test the predictions consumed by search, independently of whether the compiler reports a successful build or refit.

\subsection{Promotion from incomparable training losses}\label{sec:05a-three-failures-promotion-from-incomparable-training-losses}
The progressive controller initially used smoothed training loss as a low-cost proxy for promotion readiness, avoiding additional searched matches. Extra catch-up training, however, gave the larger candidate more optimizer quanta on the same replay distribution. Its lower loss therefore reflected unequal training exposure as well as model quality. The gate promoted a candidate approximately 270 Elo weaker than the active player, despite its deployment artifact passing inference-fidelity checks.

Promotion now requires a score of at least 0.48 in two consecutive paired matches against the active deployment artifact. A failing score resets the sequence; failed or cancelled matches do not count. This separates the optimization objective from the replacement decision: training loss guides candidate fitting, while direct play determines whether it can replace the active model. Appendix \ref{app:D} specifies candidate timing and catch-up training.

\section{Training progress under limited compute}\label{sec:06-final-chess-recipe}
The final training run produced a 6.32-million-parameter chess model in 2.5 days on one node with eight RTX 4070 SUPER GPUs and 80 logical CPUs. This chapter examines the progression in strength and the training volume made possible by the integrated system. The complete recipe is specified in Appendix \ref{app:D}; \texttt{chess-final-config.yaml} \cite{ref10} remains the entry point for reproduction.

\subsection{Progress across training campaigns}\label{sec:06-final-chess-recipe-progress-across-training-campaigns}
\begin{figure*}[t]
\centering
\includegraphics[width=494.476bp,height=0.70\textheight,keepaspectratio]{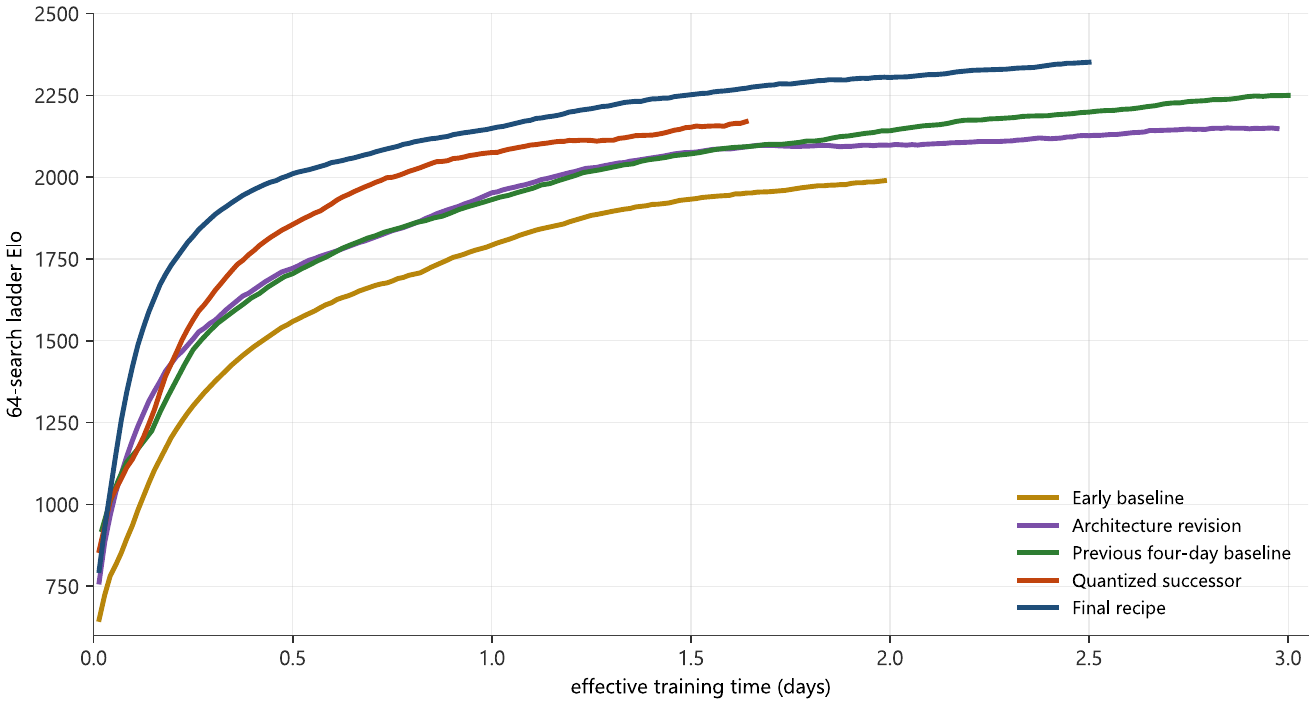}
\caption{Smoothed 64-search training ladders across five chess campaigns. The final curve ends at 2.5 days and the preceding baseline at 3 days. The deeper-search results for the final model are shown in Figure \ref{fig:final-search-curve-paper}.}\label{fig:chess-ladder-progress-paper}
\end{figure*}

Figure \ref{fig:chess-ladder-progress-paper} compares the 64-search training ladders across five chess campaigns. The final recipe reached higher playing strength within the reported training window.

\begin{figure*}[t]
\centering
\includegraphics[width=494.476bp,height=0.70\textheight,keepaspectratio]{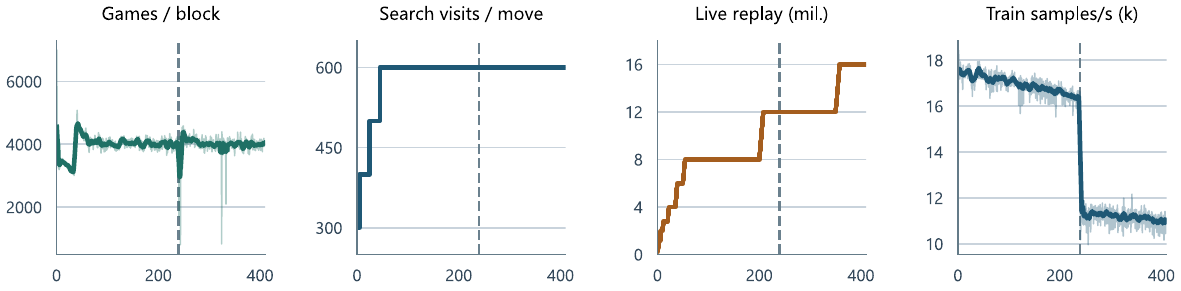}
\caption{Training volume and throughput on a common optimizer-step axis. Panels show completed games per training block, search visits per move, live replay occupancy, and trainer throughput. Replay capacity increases in steps, while the small-to-medium transition coincides with lower trainer throughput.}\label{fig:final-training-volume-and-throughput-paper}
\end{figure*}
These progress curves use only 64 searches per move to keep evaluation inexpensive during training. Their roughly 2,400-Elo level therefore reflects a much smaller thinking budget than the final model's 3,251 benchmark Elo at 100,000 searches per move. Additional MCTS search improves move selection using the trained network, without further training.

The larger progression across the plotted campaigns reflects successive changes to architecture, data generation, and training efficiency.

\subsection{Training volume and model transition}\label{sec:06-final-chess-recipe-training-volume-and-model-transition}
\begin{figure*}[t]
\centering
\includegraphics[width=494.476bp,height=0.70\textheight,keepaspectratio]{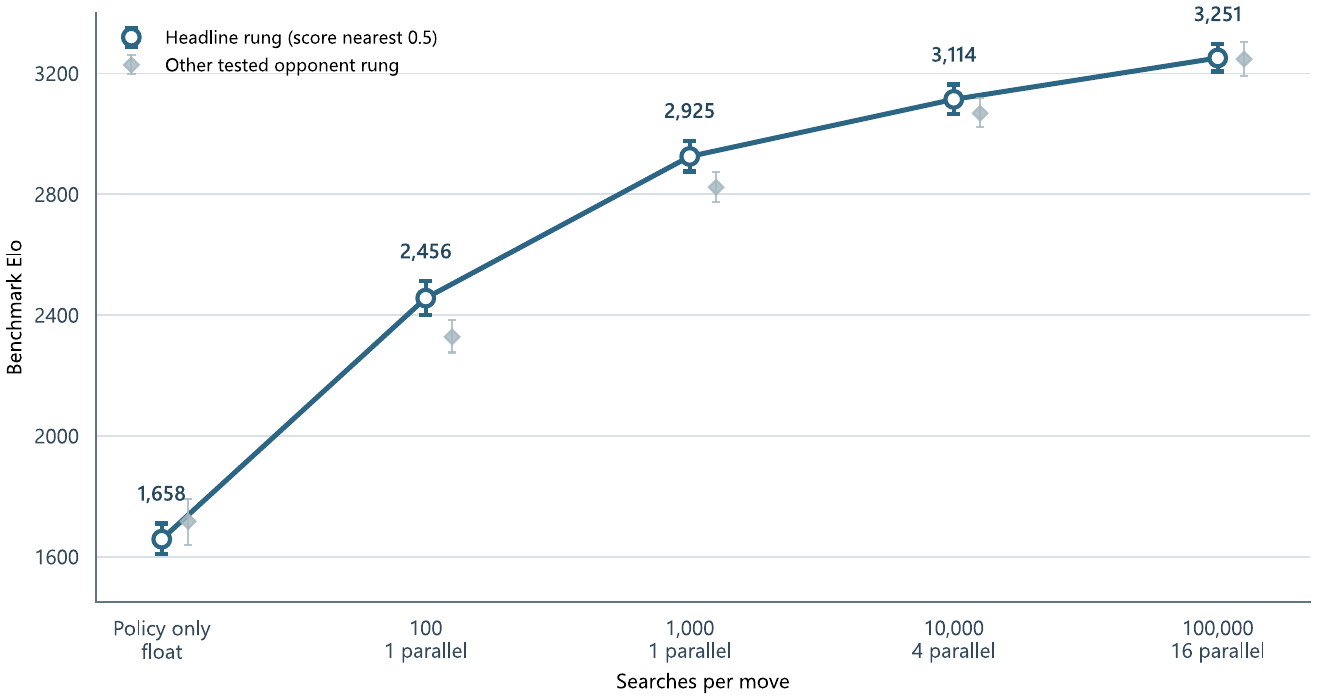}
\caption{Final playing strength across search budgets. Connected points use the opponent rung with score nearest 50\%; pale diamonds show the second opponent-based estimate and bars indicate 95\% match-bootstrap intervals. The horizontal axis lists search budgets categorically.}\label{fig:final-search-curve-paper}
\end{figure*}
Table \ref{tab:06-final-chess-recipe-1} summarizes the scale of the final run.

\begin{table}[!t]
\centering\small
\caption{Final training run}\label{tab:06-final-chess-recipe-1}
\begin{tabular}{@{}ll@{}}
\toprule
Quantity & Final training \\
\midrule
Duration & 2.5 days \\
Model parameters & 6.32 million \\
Completed games & 3.25 million \\
Mean game length & \textasciitilde{}104 plies \\
Estimated search simulations & \textasciitilde{}100 billion \\
Materialized positions & 209 million \\
Training presentations & 837 million \\
Optimizer steps & 409 thousand \\
Live replay at selection & 16.0 million positions \\
Node rental cost & \$43.2 \\
\bottomrule
\end{tabular}
\end{table}
We estimate approximately 100 billion search simulations, using 209 million materialized positions and an assumed average of 500 fresh simulations per position, allowing for subtree reuse and the lower initial visit budgets. The multiplier is an accounting assumption, not an independently measured average: 209,153,744 \ensuremath{\times} 500 gives 104,576,872,000 simulations before rounding.

Figure \ref{fig:final-training-volume-and-throughput-paper} relates this volume to replay growth and the transition from the 12\ensuremath{\times}128 to the 14\ensuremath{\times}160 model. The smaller network supported faster early training; median trainer throughput fell from about 17.0 thousand to 11.2 thousand samples/s across the two stages. Search increased from 300 to 600 visits per move over the plotted run. The configured 800-visit stage lies beyond the selected checkpoint.

The expanding replay window retained a broader history of self-play while the learner received approximately four presentations per admitted position. Throughput, search budget, and replay growth therefore changed together as training progressed. Loss, learning-rate, and gradient diagnostics are provided in Appendix \ref{app:A}.

The reported checkpoint was selected near the strongest region of the training ladder. A function-preserving 19\ensuremath{\times}176 continuation recovered parity but did not establish a higher plateau.

\section{Playing strength and model compression}\label{sec:07-final-run-results}
The final model reached \textbf{3,251 benchmark Elo at 100,000 searches per move}, estimated at under five seconds of thinking time on an RTX 4070 SUPER. This chapter examines the strength gained from search and the performance retained by a much smaller distilled model. Table \ref{tab:02-methodology-and-evidence-1} contains the full paired-match results; Appendix \ref{app:B} specifies the evaluation protocol.

\subsection{Strength across search budgets}\label{sec:07-final-run-results-strength-across-search-budgets}
Search increased playing strength from \textbf{1,658 Elo for policy-only play} to \textbf{3,251 Elo} at the deepest budget, a gain of 1,593 points. Figure \ref{fig:final-search-curve-paper} shows continuing improvement with diminishing returns: the final tenfold increase in search added 137 Elo, substantially less than the increases at shallow budgets.

The two opponent-based estimates at the deepest budget differ by only four Elo. At shallower budgets the estimates are farther apart, as shown by the paired points. Appendix \ref{app:B} describes the calibration and rating calculation.

Deeper searches use additional leaf parallelism to improve GPU utilization. The evaluation curve uses one leaf at a time at 100 and 1,000 searches, four at 10,000, and sixteen at 100,000. Section \ref{sec:04a-search} examines the strength--latency tradeoff.

\subsection{Distilling a smaller player}\label{sec:07-final-run-results-distilling-a-smaller-player}
The distilled student has approximately 470 thousand parameters, \textbf{13.4 times fewer} than the teacher. It trained for 110,000 optimizer steps on a frozen 20-million-position replay snapshot. Increasing search from 10,000 to 100,000 raised its rating from 2,697 to 2,873 Elo (Table \ref{tab:07-final-run-results-1}).

\begin{table}[!t]
\centering\small
\caption{Distilled student strength across search budgets}\label{tab:07-final-run-results-1}
\begin{tabular}{@{}lll@{}}
\toprule
Searches & Elo & 95\% interval \\
\midrule
10,000 & 2,697 & 2,640--2,753 \\
100,000 & 2,873 & 2,819--2,935 \\
\bottomrule
\end{tabular}
\end{table}
The small network retains substantial playing strength, but the gap to the teacher remains despite deeper search.

\section{Limitations and open questions}\label{sec:08-limitations}
The study demonstrates an integrated training recipe under a limited compute budget. Its principal limitations concern attribution of the gains, calibration of playing strength, and transfer to larger models or different workloads.

\subsection{Attribution and scaling}\label{sec:08-limitations-attribution-and-scaling}
Most component experiments used a single seed, a short continuation, or frozen replay. These controls supported design selection at affordable cost, but do not identify an independent final-strength contribution for each component. Policy-head and global-context comparisons also include qualitative observations rather than complete matched evaluations.

The attribution problem is strongest for coupled choices. Progressive sizing changes capacity and training history as well as self-play cost; replay growth and reuse alter both data exposure and demand for fresh games. Matched-compute experiments are needed to separate those effects.

\subsection{Persistent plateaus and the remaining engine gap}\label{sec:08-limitations-persistent-plateaus-and-the-remaining-engine-gap}
Later continuation experiments showed diminishing strength gains, with similar plateaus at 64 and 400 searches per move. Appendix \ref{sec:appendix-c-supporting-comparisons-late-training-strength} presents the checkpoint comparisons and late-training ladder observations. Further training may yield incremental gains on the order of tens of Elo, but these trajectories do not suggest improvements of hundreds of points. Closing the roughly 400-Elo gap to the upper end of the historical Stockfish 13 calibration would likely require changes to the training recipe.

Four changes are the most plausible next tests: substantially larger models to increase representational capacity; more search during self-play to improve targets; lower replay reuse to increase fresh-game exposure; and a larger replay window to retain more diverse experience. Each changes a different potential limit on learning, and each also changes compute demand or data age. They should be compared by strength gained at matched compute, not solely by training loss or update count.

\subsection{Rating calibration and evaluation scope}\label{sec:08-limitations-rating-calibration-and-evaluation-scope}
The benchmark scale derives from a published calibration of fixed-node Stockfish 13 \cite{ref9}, rather than direct matches against unrestricted engines. Each final match contains 100 games, and the reported intervals capture match sampling uncertainty with the calibration anchors held fixed. Agreement between opponent rungs is close at the deepest budget and weaker at shallow budgets. A larger match set and broader opponent field would help distinguish sampling variation from calibration and matchup effects.

The search-budget curve combines deeper search with increased parallelism. It measures the deployed operating points, not the isolated effect of search depth. Backend differences likewise enter policy-only, searched, and student evaluations.

\subsection{Workload dependence and development cost}\label{sec:08-limitations-workload-dependence-and-development-cost}
The rejected optimizations remain plausible under other workloads. Diverse chess self-play produced too few exact graph or inference-cache hits to offset lookup and synchronization costs; repeated analysis may offer more reuse. In the GPU-bound training workload, actor-trainer overlap limited the cadence gain from adaptive stopping: reducing search work did not proportionally reduce the complete training cycle.

The \$43.20 rental cost covers the final training run, excluding development experiments and evaluation. It characterizes the cost of executing the recipe rather than discovering it. Reproduction also depends on hardware contention, drivers, and TensorRT versions; the configuration, published model, and evaluation protocol define the comparison, while exact weights and timing may vary across executions.

\section{Conclusion}\label{sec:10-conclusion}
In 2.5 days on one eight-GPU node, an AlphaZero-style chess system trained from random initialization produced a 6.3-million-parameter model measuring \textbf{3,251 benchmark Elo at 100,000 searches per move} against a fixed-node Stockfish 13 ladder. Without search, the same model measured \textbf{1,658 benchmark Elo}.

The system combines a compact structured policy head and shared convolutional backbone with progressive small-to-medium sizing, targeted restart states, and prioritized replay. An INT8-capable architecture, native search, batched TensorRT inference, and distributed training supplied the required volume of self-play data and optimizer updates. The result demonstrates the strength achievable by this integrated recipe within the available compute budget.

The component studies also identify limits to local optimization. Better deep-policy fidelity from learned search allocation did not produce better online learning, and fewer simulations from adaptive stopping yielded little reduction in cycle time. Exact graph and neural-cache reuse did not offset their overhead. More seriously, removing searched endgames corrupted value targets, successful TensorRT refits altered predictions, and loss-based promotion selected a weaker player. These findings connect efficiency to the quality of the data and model actually consumed by self-play.

Further scaling remains open. The larger model recovered its parent's strength without surpassing it during the available continuation, while distillation retained substantial strength in a much smaller network without matching the teacher. Continued training may yield modest gains, but the observed plateaus point toward changes to the recipe for a substantial improvement. Larger models, deeper self-play search, lower reuse, and broader replay are the next hypotheses to test against measured strength per unit of compute.

\FloatBarrier
\balance
\section*{Acknowledgements}
Generative AI tools assisted with manuscript drafting, revision, and figure tooling. The author reviewed the resulting material and takes responsibility for the final manuscript.

\clearpage\onecolumn\raggedbottom\widowpenalty=150\clubpenalty=150
\appendix
\renewcommand{\thefigure}{\Alph{section}.\arabic{figure}}
\renewcommand{\thetable}{\Alph{section}.\arabic{table}}
\Needspace{6\baselineskip}\setcounter{figure}{0}\setcounter{table}{0}
\section{Training diagnostics}\label{app:A}
The diagnostic traces cover the training sequence that produced checkpoint 1026: 817 quanta of 500 optimizer steps at global batch 2,048, totaling 408,500 steps and 836,608,000 presentations. Checkpoint identifiers include additional experimental branches, whereas these plots exclude the reverted INT8 conversion, invalid capacity promotion, and subsequent larger-model continuation. Checkpoint 1026 was selected near the strongest 64-search ladder region and was the last checkpoint in that region with the complete model, optimizer, QAT, ONNX, and TensorRT artifacts retained.

\begin{figure}[H]
\centering
\includegraphics[width=431.527bp,height=0.70\textheight,keepaspectratio]{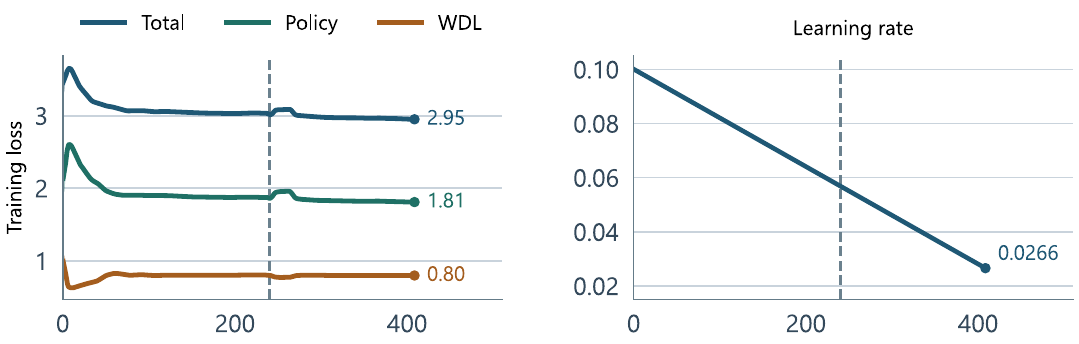}
\caption{Training losses and learning rate. Raw losses are overlaid with 11-quantum moving averages; the dashed line marks the small-to-medium transition at 240,000 optimizer steps. Total loss reaches approximately 2.95 while WDL loss remains near 0.80. The learning rate declines from approximately 0.10 to 0.0266. Labels identify the final values.}\label{fig:final-training-loss-and-rate-paper}
\end{figure}
The exact counters underlying Chapter \ref{sec:06-final-chess-recipe} are 3,249,647 completed games and approximately 209.15 million net materialized positions, with 16 million live rows at selection. Median trainer throughput was 16,977 samples/s over 480 small-model quanta and 11,194 over 337 medium-model quanta. Figure \ref{fig:final-training-volume-and-throughput-paper} in Chapter \ref{sec:06-final-chess-recipe} shows these trajectories. Node rental was \$0.72/hour, totaling \$43.20 for the 2.5-day run.

\begin{figure}[H]
\centering
\includegraphics[width=406.985bp,height=0.70\textheight,keepaspectratio]{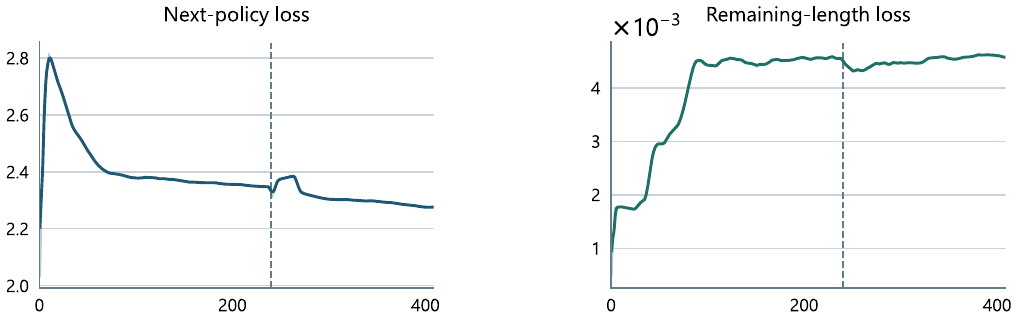}
\caption{Next-policy and remaining-game-length losses on separate scales. Faint traces show recorded values; solid curves show 11-quantum moving averages. The dashed line marks the small-to-medium transition, as in the remaining optimizer-step plots.}\label{fig:appendix-auxiliary-losses}
\end{figure}
\begin{figure}[H]
\centering
\includegraphics[width=349.291bp,height=0.70\textheight,keepaspectratio]{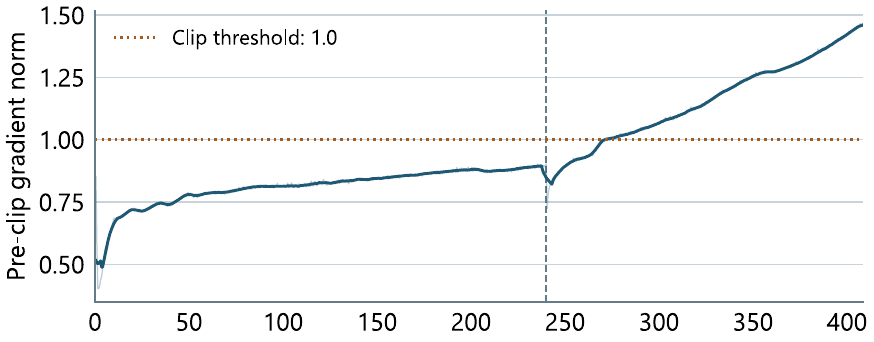}
\caption{Mean pre-clipping gradient norm in each 500-step training block. The dotted line is the configured norm limit of 1.0. The mean rises above that limit during medium-model training; the optimizer clips individual steps before applying them.}\label{fig:appendix-gradient-norm}
\end{figure}
\begin{figure}[H]
\centering
\includegraphics[width=361.134bp,height=0.70\textheight,keepaspectratio]{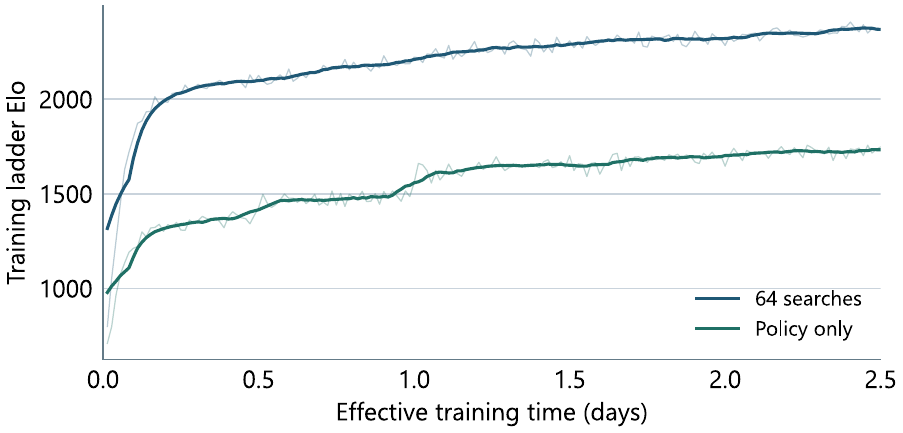}
\caption{Policy-only and 64-search training-ladder ratings over the final 2.5-day window. Faint traces show individual evaluations and solid curves average eleven evaluations. Both improve over training; Table \ref{tab:02-methodology-and-evidence-1} reports the separate final matches.}\label{fig:appendix-policy-search-progress}
\end{figure}
Replay occupancy, sampled-position age, and resignation calibration are shown with their corresponding methods in Section \ref{sec:04b-data-and-replay} (Figure \ref{fig:appendix-replay-age} and Figure \ref{fig:appendix-resignation}).

\Needspace{6\baselineskip}\setcounter{figure}{0}\setcounter{table}{0}
\section{Evaluation detail}\label{app:B}
This appendix specifies the match protocol and rating calculation underlying the results in Chapters 2 and 8.

\subsection{Fixed-node match protocol}\label{sec:appendix-b-evaluation-tables-fixed-node-match-protocol}
Each of the ten teacher matches in Table \ref{tab:02-methodology-and-evidence-1} comprises 100 games from 50 openings played with colours reversed. The opening set is \texttt{chess-stockfish-8moves-v3-openings-v33.json} (SHA-256 prefix \texttt{490425ed0f466f55}). Each opening supplies eight full moves (16 plies) before evaluation play. Games reaching the 300-ply evaluation cap are recorded as draws and included in the fixed-opponent score. Stockfish 13 used one thread and 1,024 MiB hash. Its fixed node limits were assigned the historical benchmark-Elo anchors of 1,700 at 1,000 nodes, 1,890 at 2,000, 2,220 at 5,000, 2,470 at 10,000, 2,700 at 20,000, 2,960 at 50,000, 3,100 at 100,000, and 3,230 at 200,000 \cite{ref9}. These anchors were read from the published calibration curve. The reported rating at each candidate budget uses the tested opponent against which the score is closest to 0.5. Searched games used the INT8 TensorRT artifact; policy-only games used the corresponding float TorchScript export.

Evaluation uses an exploration constant of 1.0, no root noise, and deterministic maximum-visit move selection; policy-only play selects the highest-probability legal action. Each inference queue uses one worker, batch capacity 64, and one outstanding batch. Teacher leaf parallelism is specified in Table \ref{tab:02-methodology-and-evidence-1}.

For score \texttt{s} against an opponent anchor \texttt{R}, benchmark Elo is \texttt{R + 400 log10(s / (1 - s))}. Each 95\% interval resamples the 50 colour-swapped opening pairs 10,000 times, then transforms the 2.5th and 97.5th percentiles of the resulting match scores with the anchor held fixed. The intervals therefore quantify match sampling uncertainty, excluding uncertainty in the historical calibration.

At 100, 1,000, 10,000, and 100,000 searches, the harder opponent implies a rating 128, 102, 46, and 4 Elo higher than the easier opponent. The discrepancy narrows with search budget, reaching close agreement at the deepest point. Draw behaviour, calibration error, matchup effects, and sampling variation could contribute; the matches do not distinguish them. Selecting the score nearest 0.5 limits extrapolation from lopsided results.

\subsection{Student matches}\label{sec:appendix-b-evaluation-tables-student-matches}
The final student uses TorchScript and four parallel leaf searches. Both reported matches use the same 20,000-node Stockfish opponent, anchored at 2,700 benchmark Elo, with 100 games each. At 10,000 searches its W/D/L count is 35/29/36, yielding 2,697 [2,640, 2,753] Elo. At 100,000 searches it scores 59/28/13, yielding 2,873 [2,819, 2,935] Elo. Intervals use the paired bootstrap described above. The deepest student result used one opponent rung.

\subsection{Thinking-time estimate}\label{sec:appendix-b-evaluation-tables-thinking-time-estimate}
The under-five-second estimate for 100,000 searches on an RTX 4070 SUPER extrapolates from approximately five seconds for 80,000 searches with TorchScript and approximately 1.8\ensuremath{\times} faster inference with TensorRT. Proportional scaling gives approximately 3.5 seconds:

\begin{equation}
5\,\mathrm{s}\times\frac{100{,}000}{80{,}000}\div 1.8 \approx 3.5\,\mathrm{s}.
\end{equation}
This assumes inference acceleration transfers sufficiently to overall search; 3.5 seconds is not a measured latency. The earlier batched timing is a throughput-derived estimate rather than an isolated single-game test.

\Needspace{6\baselineskip}\setcounter{figure}{0}\setcounter{table}{0}
\section{Supporting search and systems comparisons}\label{app:C}
The following controls provide the numerical comparisons supporting the search and systems decisions in Chapters 4 and 5. Each comparison retains its own workload and measurement scale.

\subsection{Architecture shape and inference throughput}\label{sec:appendix-c-supporting-comparisons-architecture-shape-and-inference-throughput}
The width sweep used 12-block global-pooling CNNs with a dense policy head, BF16, batch 512, and one RTX 4070 SUPER. Each width was benchmarked in a separate process against an interleaved width-128 reference. Table \ref{tab:appendix-c-supporting-comparisons-1} contrasts measured throughput with the inverse-width-squared estimate implied by convolutional arithmetic. Ratios below are normalized to the unrounded width-128 rate of 103,490 positions/s.

\begin{table}[H]
\centering\normalsize
\caption{CNN width: measured throughput versus arithmetic prediction}\label{tab:appendix-c-supporting-comparisons-1}
\setlength{\tabcolsep}{9pt}
\renewcommand{\arraystretch}{1.1}
\begin{tabular}{@{}llll@{}}
\toprule
Depth \ensuremath{\times} width & Positions/s (k) & Measured ratio & Arithmetic ratio \\
\midrule
12\ensuremath{\times}96 & 133 & 1.284 & 1.778 \\
12\ensuremath{\times}112 & 95.9 & 0.926 & 1.306 \\
12\ensuremath{\times}120 & 93.4 & 0.903 & 1.138 \\
12\ensuremath{\times}128 & 103 & 1.000 & 1.000 \\
12\ensuremath{\times}136 & 59.7 & 0.577 & 0.886 \\
12\ensuremath{\times}144 & 63.8 & 0.617 & 0.790 \\
12\ensuremath{\times}152 & 53.8 & 0.520 & 0.709 \\
12\ensuremath{\times}160 & 61.8 & 0.597 & 0.640 \\
12\ensuremath{\times}176 & 52.4 & 0.507 & 0.529 \\
12\ensuremath{\times}192 & 41.5 & 0.401 & 0.444 \\
12\ensuremath{\times}224 & 35.5 & 0.343 & 0.327 \\
12\ensuremath{\times}256 & 31.1 & 0.301 & 0.250 \\
\bottomrule
\end{tabular}
\end{table}
The 112- and 120-channel models perform less arithmetic than the 128-channel model but serve fewer positions per second. The sharp loss at 136 channels likewise exceeds the arithmetic prediction. Width alone is therefore an unreliable proxy for inference cost.

Table \ref{tab:appendix-c-supporting-comparisons-2} extends the comparison to depth and serving batch size. Ratios are relative to the specified reference at the same batch size. The width sweep is backed by per-process measurements; the depth/batch comparisons are transcribed benchmark summaries. Dashes indicate settings not measured.

\begin{table}[H]
\centering\normalsize
\caption{Depth and width: throughput ratios across batch sizes}\label{tab:appendix-c-supporting-comparisons-2}
\setlength{\tabcolsep}{9pt}
\renewcommand{\arraystretch}{1.1}
\begin{tabular}{@{}lllll@{}}
\toprule
Model & Reference & Batch 512 & Batch 320 & Batch 64 \\
\midrule
20\ensuremath{\times}128 & 14\ensuremath{\times}152 & 1.36 & 1.20 & 0.730 \\
10\ensuremath{\times}176 & 14\ensuremath{\times}152 & 1.35 & 1.24 & 1.34 \\
13\ensuremath{\times}160 & 14\ensuremath{\times}152 & 1.23 & 1.13 & 1.10 \\
14\ensuremath{\times}160 & 14\ensuremath{\times}152 & -- & 1.05 & 1.01 \\
15\ensuremath{\times}160 & 14\ensuremath{\times}152 & -- & 0.980 & 0.930 \\
16\ensuremath{\times}160 & 14\ensuremath{\times}152 & -- & 0.918 & 0.960 \\
14\ensuremath{\times}176 & 14\ensuremath{\times}152 & -- & 0.888 & 0.956 \\
11\ensuremath{\times}224 & 18\ensuremath{\times}176 & 1.10 & 1.24 & 1.66 \\
34\ensuremath{\times}128 & 18\ensuremath{\times}176 & 1.06 & 1.16 & 0.556 \\
19\ensuremath{\times}176 & 18\ensuremath{\times}176 & -- & 0.948 & 0.970 \\
20\ensuremath{\times}176 & 18\ensuremath{\times}176 & -- & 0.899 & 0.912 \\
22\ensuremath{\times}160 & 18\ensuremath{\times}176 & -- & 0.995 & 0.838 \\
4\ensuremath{\times}224 & 12\ensuremath{\times}128 & 0.978 & 1.09 & 2.36 \\
6\ensuremath{\times}176 & 12\ensuremath{\times}128 & 0.977 & 1.01 & 1.74 \\
\bottomrule
\end{tabular}
\end{table}
The 20\ensuremath{\times}128 network is faster than 14\ensuremath{\times}152 at batch 512 but slower at batch 64. Such reversals motivate measuring candidate models at both self-play and interactive batch sizes rather than extrapolating from parameter count.

The attention comparison controlled bootstrap policy shape, serving precision, and runtime. Its best Smolgen cell reduced held-out loss by 0.0090 nats relative to the parameter-matched from-to CNN, but delivered only 36.1\% of the dense CNN reference's batch-512 forward rate and 45.8\% at batch 64, with 5.17\ensuremath{\times} its peak training memory. These rates use the dense CNN reference, not the parameter-matched from-to comparison.

\subsection{Replay-reuse controls}\label{sec:appendix-c-supporting-comparisons-replay-reuse-controls}
Approximately 90-minute controls at reuse ratios 4, 6.25, and 8 showed comparable strength at their shared evaluation boundaries despite different update rates. The retained ratio of four favours fresh self-play data. Its long-run effect is coupled to replay capacity, optimization, and inference changes in the completed campaign. Reuse also changes the wall-clock pace of evaluation, publication, search-budget schedules, and replay growth, which advance at quantum boundaries. The configured credit per admitted row therefore describes the training schedule rather than the exact exposure of every distinct replay position.

\subsection{Negative search and reuse controls}\label{sec:appendix-c-supporting-comparisons-negative-search-and-reuse-controls}
The learned pre-search allocator captured approximately 23\% of the available improvement in deep-policy divergence at nearly matched search cost, but trailed non-adaptive online training by roughly 60--100 ladder Elo. The in-search stopper skipped approximately 14\% of nominal simulations while improving generation cadence by only about 3\% under actor-trainer overlap. Its paired strength differences, calculated as baseline minus stopper (+1.7 \ensuremath{\pm} 9.9 and \ensuremath{-}4.2 \ensuremath{\pm} 10.1 Elo, standard errors), did not resolve an effect.

Exact graph search avoided only 0.0249\% and 0.1769\% of neural evaluations at 1,000 and 10,000 searches while being 8.63\% and 8.28\% slower. A bounded inference cache shared inside each self-play process had a 0.970\% hit rate; disabling it made game updates about 0.88\% faster and lowered peak worker memory about 7.12\%. A wider unbounded repeat tracker observed at most about 4\% reuse in the tested production-like workloads and itself cost throughput. In these workloads, the measured reuse was insufficient to offset the cost of detecting and exploiting it.

\subsection{Throughput benchmarks}\label{sec:appendix-c-supporting-comparisons-throughput-benchmarks}
\textbf{Host submission.} The controlled 32-process self-play benchmark increased aggregate search throughput from 512,679 to 617,782 searches/s. On the same node restricted to 24 CPU cores, it increased from 215,265 to 496,036 searches/s. These are rates across the actor workload, including search and inference.

\textbf{Neural inference.} The matched runtime benchmark measured 75,889 positions/s with TensorRT FP16 and 40,716 with TorchScript BF16. These count positions evaluated by the network, not aggregate self-play searches.

\textbf{Replay delivery.} Compacting 5,000 producer shards into 25 containers increased loader throughput on 2.5 million rows from 3,943 to 32,579 samples/s. In a separate live interval, materialization appended 8,790 positions/s against an arrival rate of 1,365 accepted positions/s.

\textbf{Training.} An eight-GPU benchmark with global batch 2,048, bfloat16 autocast, and concurrent self-play processed 6,252 training samples/s. The actor-overlap sweep in Table \ref{tab:05-systems-optimization-1} is a separate workload; its trainer and concurrent-search rates should be compared within that sweep.

Table \ref{tab:appendix-c-supporting-comparisons-3} retains the unrounded trainer measurements. Chapter \ref{sec:05-systems-optimization} presents the comparison alongside the scheduling decision. Cycle times are estimated from measured training duration and search throughput.

\begin{table}[H]
\centering\normalsize
\caption{Trainer throughput under actor overlap}\label{tab:appendix-c-supporting-comparisons-3}
\setlength{\tabcolsep}{9pt}
\renewcommand{\arraystretch}{1.1}
\begin{tabular}{@{}lll@{}}
\toprule
Active actors & Training samples/s & Estimated cycle time (s) \\
\midrule
0 & 25,275 & -- \\
8 & 21,492 & 117.1 \\
16 & 17,139 & 112.9 \\
32 & 9,210 & 111.2 \\
\bottomrule
\end{tabular}
\end{table}
The 1.86x TensorRT INT8 versus TorchScript search comparison used a production-shaped 400-visit actor workload, with different checkpoint weights, so the comparison includes both backend and model changes. The 23.3\% replay-admission gain compared separate live 400- and 600-visit stages that also differed in checkpoint and actor scheduling. Absolute systems rates vary with CPU quota, GPU power, PCIe and NUMA layout, runtime versions, model shape, batching, and concurrent training.

\subsection{Late-training strength}\label{sec:appendix-c-supporting-comparisons-late-training-strength}
A retrospective comparison evaluated checkpoints 900, 960, and 1020 at 64 and 400 searches per move (Table \ref{tab:appendix-c-supporting-comparisons-4}). Each cell used 50 paired openings, giving 100 games. The 64-search matches used Stockfish 13 at 10,000 nodes (2,470 benchmark Elo); the 400-search matches used 20,000 nodes (2,700 benchmark Elo). Ratings are calculated from each match score against its opponent anchor.

\begin{table}[H]
\centering\normalsize
\caption{Late-training checkpoint comparisons at two search budgets}\label{tab:appendix-c-supporting-comparisons-4}
\setlength{\tabcolsep}{9pt}
\renewcommand{\arraystretch}{1.1}
\begin{tabular}{@{}lllll@{}}
\toprule
Checkpoint & Searches & W/D/L & Score & Benchmark Elo \\
\midrule
900 & 64 & 18/26/56 & 31.0\% & 2,331.0 \\
960 & 64 & 16/31/53 & 31.5\% & 2,335.0 \\
1020 & 64 & 16/34/50 & 33.0\% & 2,347.0 \\
900 & 400 & 31/41/28 & 51.5\% & 2,710.4 \\
960 & 400 & 33/38/29 & 52.0\% & 2,713.9 \\
1020 & 400 & 26/36/38 & 44.0\% & 2,658.1 \\
\bottomrule
\end{tabular}
\end{table}
The 64-search estimate increased by 16.0 Elo between the first and last checkpoints, while the 400-search matches showed no sustained gain. Separately, a late continuation produced 28 successive observations on the three-rung 64-search ladder over 9.2 hours. The mean of the first 14 observations was 2,373.7 Elo, compared with 2,376.7 for the last 14: a change of 3.0 Elo. These within-protocol comparisons support diminishing returns from continued training, rather than demonstrating that further improvement is impossible.

\Needspace{6\baselineskip}\setcounter{figure}{0}\setcounter{table}{0}
\section{Reproducibility and release boundary}\label{app:D}
This appendix specifies the network representation, numerical recipe, and artifacts required to reproduce the software and evaluation. The released configuration supports new training runs; the published checkpoint fixes the model used for the reported results.

\subsection{Network input and output contract}\label{sec:appendix-d-reproducibility-network-input-and-output-contract}
The network input is a 52\ensuremath{\times}8\ensuremath{\times}8 tensor with channels listed in Table \ref{tab:appendix-d-reproducibility-1}. "Own" denotes the player to move. For Black to move, ranks are reflected and piece colours are exchanged; files retain their order. Tensor row zero is the player's home rank, and column zero is file a. There is no separate absolute-colour plane. Spatial masks contain zeros and ones; flags and scalar values fill all 64 squares.

\begin{table}[H]
\centering\normalsize
\caption{Chess input planes in tensor order (zero-based indices)}\label{tab:appendix-d-reproducibility-1}
\setlength{\tabcolsep}{9pt}
\renewcommand{\arraystretch}{1.1}
\begin{tabularx}{\textwidth}{@{}l X l@{}}
\toprule
Plane indices & Feature, in channel order & Encoding \\
\midrule
0, 1, 2, 3, 4, 5 & Own pawn, knight, bishop, rook, queen, king & Piece-location masks \\
6, 7, 8, 9, 10, 11 & Opponent pawn, knight, bishop, rook, queen, king & Piece-location masks \\
12 & Own kingside castling right & Constant 0 or 1 \\
13 & Own queenside castling right & Constant 0 or 1 \\
14 & Opponent kingside castling right & Constant 0 or 1 \\
15 & Opponent queenside castling right & Constant 0 or 1 \\
16 & All own pieces & Occupancy mask \\
17 & All opponent pieces & Occupancy mask \\
18 & Pieces checking the player to move & Checker-location mask \\
19 & En-passant target & One square, or all zeros \\
20 & At least one earlier occurrence of this position & Constant 0 or 1 \\
21 & At least two earlier occurrences of this position & Constant 0 or 1 \\
22, 23 & Most recent move: origin, destination & Two single-square masks \\
24, 25 & Second-most-recent move: origin, destination & Two single-square masks \\
26, 27 & Third-most-recent move: origin, destination & Two single-square masks \\
28, 29 & Fourth-most-recent move: origin, destination & Two single-square masks \\
30, 31 & Fifth-most-recent move: origin, destination & Two single-square masks \\
32, 33 & Sixth-most-recent move: origin, destination & Two single-square masks \\
34, 35 & Seventh-most-recent move: origin, destination & Two single-square masks \\
36, 37 & Eighth-most-recent move: origin, destination & Two single-square masks \\
38 & Fixed checkerboard, with a1 set to one & Alternating 0/1 mask \\
39 & Exactly one bishop per side, on opposite colours & Constant 0 or 1 \\
40, 41, 42, 43, 44, 45 & Pawn, knight, bishop, rook, queen, king balance & Own count minus opponent count \\
46 & Halfmove clock & Integer count, capped at 100 \\
47, 48, 49, 50, 51 & Own pawn, knight, bishop, rook, queen counts & Integer counts \\
\bottomrule
\end{tabularx}
\end{table}
Channels 0--39 are binary; channels 40--51 contain unnormalized scalar values. Missing history entries are zero-filled. Move history stores origin and destination pairs for the last eight plies rather than complete board states; castling records the king's actual destination. The checkerboard is fixed in canonical coordinates.

The shared backbone produces 160\ensuremath{\times}8\ensuremath{\times}8 features. The policy head projects each square to a 128-dimensional token and forms query and key vectors whose scaled dot products score origin-destination pairs. These scores are gathered into 1,880 action logits: 1,792 ray or knight pairs and 88 explicit promotion actions. Promotion offsets distinguish queen, rook, bishop, and knight choices. Castling uses the king-to-own-rook pair in the action encoding, while en passant uses the pawn's ordinary origin-destination pair. Illegal actions are masked before softmax over the legal moves.

The value branch uses a two-channel 1\ensuremath{\times}1 convolution, batch normalization, ReLU, flattening, a 48-unit hidden layer, and three logits. Softmax produces win, draw, and loss probabilities for the player to move. The auxiliary branches produce a further 1,880 logits for the next searched policy and one scalar for remaining game length. They are used only in training; inference returns the primary policy logits and WDL probabilities.

\subsection{Configuration and reported result}\label{sec:appendix-d-reproducibility-configuration-and-reported-result}
The fully expanded \texttt{chess-final-config.yaml} is the maintained entry point for training with the current recipe. Replication of the reported experiment instead uses its frozen source revision, resolved configuration, manifest, checkpoints, engines, and datasets. This separates future recipe updates from the fixed experimental record.

\subsection{Expanded chess recipe settings}\label{sec:appendix-d-reproducibility-expanded-chess-recipe-settings}
The final recipe ran on one node with eight RTX 4070 SUPER GPUs and 80 logical CPUs. Its network ladder contains 12\ensuremath{\times}128, 14\ensuremath{\times}160, and 19\ensuremath{\times}176 residual CNNs with capped scaled post-activation branches and global-pooling context in every second block. The reported checkpoint uses the 14\ensuremath{\times}160 stage.

The architecture uses a key-size-128 chess from-to policy head and a two-channel WDL head with a 48-unit hidden layer. The training-only next-searched-policy and remaining-game-length heads have loss weights 0.15 and 0.1. Primary policy and value losses each have weight 1.0. Terminal outcome targets are discounted by 0.998 per ply; the search-root-value blend rises from zero to 0.1 over its configured schedule, while search backup uses a separate 0.99 per-ply discount. At a capped game's final position, a searched scalar value \texttt{v} is converted to a soft WDL target with \texttt{r = 1 - |v|}: win, draw, and loss receive \texttt{max(v, 0) + r/3}, \texttt{r/3}, and \texttt{max(-v, 0) + r/3}, respectively. Capped games omit the remaining-game-length loss. Batch-normalization folding, recalibration, ONNX export, and TensorRT refitting operate on a deployment copy, leaving the trainable model unfused.

Each 500-step training quantum uses eight ranks processing 256 positions each, for a global batch of 2,048. Nesterov SGD uses momentum 0.9, weight decay 0.0001, and gradient clipping at norm 1.0. The learning rate warms from zero to 0.1 over the first 1,000 optimizer steps and then follows the configured linear schedule to 0.01. Training uses bfloat16 and persistent trainer processes; \texttt{torch.compile} is disabled. QAT calibration uses 516 real evaluation positions and is refreshed at every publication boundary.

The 32 self-play actors run four per GPU, with 512 interleaved games per actor. Native inference uses batches of 320 with two outstanding batches per worker. Search uses exploration constant 1.5, reduced-parent FPU with reduction 0.2, forced playout coefficient 1.5, and Dirichlet epsilon 0.25 with alpha 0.3. Eight materializers convert completed games into the fixed-layout memory-mapped replay store; training credit is committed only against durable admitted rows. Restart positions require at least 15 plies remaining in the source game, absolute root value at most 0.8, and two or three leading actions covering 85\% of visit mass. The played branch is marked used, and reservations keep workers from claiming the same alternative concurrently. Replay's logical capacity grows through 0.6, 1.2, 2.0, 2.8, 4, 6, 8, 12, 16, and 20 million rows. Restart selection is 30\% uniform and otherwise weighted by the square root of value correction, defined as half the absolute difference between searched root value and raw network value. Age and capacity bounds remove old states, and the archive is local to each worker.

The successor network trains on a captured replay snapshot for an average of 1.5 optimizer quanta per active-model quantum, with its own catch-up learning-rate clock. The configured match gate requires the successor to score at least 0.48 in two consecutive paired evaluations. The fully expanded config \cite{ref10} specifies the stage-specific ladder-plateau thresholds and window.

\subsection{Self-play schedules and promotion}\label{sec:appendix-d-reproducibility-self-play-schedules-and-promotion}
Half the actors pause during each training quantum. The search schedule rises from 300 to 800 visits per move; the reported checkpoint lies in the 600-visit stage. Games start approximately equally often from random legal openings of up to eight plies and from eligible restart states, falling back to a random opening when no restart is available. Game caps increase from 150 to 250 plies, and greedy move selection starts later as training matures. Twenty percent of games are designated as no-resignation continuations to calibrate the resignation threshold against a 2.5\% upper-bound target for mistakenly resigning a win or draw.

Each admitted replay position funds four training presentations. Sampling assigns 30\% uniform probability and otherwise prioritizes bounded policy surprise. Candidate training begins when smoothed 64-search ladder gains fall below 15 Elo/hour for the small stage or 4 Elo/hour for the medium stage. Promotion uses the paired-match gate specified above. Training monitors run every 20 minutes using 50 paired openings, a three-rung adaptive Stockfish 13 bracket, and both 64-search and policy-only play.

\Needspace{8\baselineskip}\subsection{Result identity and provenance}\label{sec:appendix-d-reproducibility-result-identity-and-provenance}
The experiment archive records the following provenance:

\begin{itemize}
\item Git revision and clean/dirty state, resolved YAML and SHA-256, and dependency-lock hash;

\item operating image, software and driver versions, GPU, CPU, RAM, disk, and engine versions and archive hashes;

\item evaluation dataset and opening-suite identities, hashes, raw matches, aggregates, commands, and interval method;

\item run manifest, approval, coordinator logs, TensorBoard events, and resource telemetry;

\item replay schema, capacity, occupancy, and reconciled volume counters;

\item checkpoint and inference hashes, ONNX and TensorRT provenance, calibration positions, and fidelity reports;

\item one digest covering the fetched archive or a checksummed artifact manifest.

\end{itemize}
The published INT8 ONNX has SHA-256 \texttt{d634abacae3c874eac6ded89f6af861eb81b509da638b5ad710587b1a08be658} at the immutable model-repository revision \texttt{dc8fccccb67ab5ec9e36267a165a9700b7dbf55f} \cite{ref11}. The source release's evidence index \cite{ref10} records additional artifact hashes. The frozen training-source revision, resolved configuration, large run archives, and some exact evaluation inputs remain in the local result archive. The public recipe and model support inspection and a new run, but are not yet a self-contained package for exact-match reproduction of the reported experiment.

\subsection{Reproducing the software}\label{sec:appendix-d-reproducibility-reproducing-the-software}
The public source release \cite{ref10} contains setup and validation instructions. Node setup uses \path{deployment/setup_remote.sh} to install locked dependencies, build the Release extension, install pinned engines, and verify their execution. Run lifecycle management uses \path{deployment/run_control.sh}.

Original project code and documentation, including this report, are available under the MIT License in the source release \cite{ref10}. The published final model artifacts carry the same license in the model repository \cite{ref11}. External dependencies, reference sources, and third-party data retain their own terms.

\subsection{Reproducing evaluation}\label{sec:appendix-d-reproducibility-reproducing-evaluation}
Matched evaluation uses the published checkpoint and inference artifact, paired opening suite, opponent binary, node limits, thread and hash settings, candidate search budget, parallelism, batching, and adjudication rules. Re-exporting with a different toolchain can change predictions and constitutes a new deployment comparison. Appendix \ref{app:B} specifies the final match protocol and rating calculation.

Performance measurements distinguish isolated neural-network throughput, saturated multi-game search, and single-game latency. These measure different execution regimes and require separate benchmarks.

\subsection{Reproducing plots and tables}\label{sec:appendix-d-reproducibility-reproducing-plots-and-tables}
Plots and tables are generated from archived JSON, CSV, TensorBoard, and manifest data, with extraction methods retained alongside the evidence. The underlying columns support recomputation of totals, rates, Elo estimates, and uncertainty intervals independently of the live dashboards.

\end{document}